\documentclass[journal]{IEEEtran}

\usepackage{cite}

\ifCLASSINFOpdf
  \usepackage[pdftex]{graphicx}
\else
\fi
\usepackage{amsmath}
\usepackage[switch]{lineno}

\usepackage{array}
\usepackage{url}

\usepackage{xspace}
\usepackage{threeparttable}
\usepackage{multirow}
\usepackage{caption}
\usepackage{subcaption}
\usepackage{wrapfig}
\usepackage{booktabs}
\usepackage{pifont}
\usepackage[colorlinks=true, linkcolor=blue, citecolor=blue, urlcolor=blue]{hyperref}
\usepackage{cleveref}
\usepackage{makecell}
\usepackage{enumitem}
\usepackage{marvosym}

\newcommand{\textmask}{semantic descriptors\xspace}
\newcommand{\rle}{R-RLE\xspace}
\newcommand{\name}{Text4Seg\xspace}
\newcommand{\namexs}{Text4Seg}
\newcommand{\namepp}{Text4Seg++\xspace}
\usepackage{tcolorbox}
\usepackage{xcolor,colortbl}
\colorlet{lightpink}{pink!35}
\colorlet{lightcyan}{cyan!20}
\colorlet{lightgray}{gray!15}
\definecolor{darkgray}{rgb}{0.9, 0.9, 0.9}
\definecolor{lightgreen}{rgb}{0.886, 0.941, 0.851}

\newcommand{\second}[1]{\textcolor{blue}{\underline{#1}}}
\newcommand{\best}[1]{\textcolor{red}{\textbf{#1}}}
\newcommand{\eg}{\emph{e.g.}}
\newcommand{\ie}{\emph{i.e.}}

\begin{document}
%
\title{Text4Seg++: Advancing Image Segmentation via Generative Language Modeling}
%
%
%

\author{
Mengcheng~Lan, Chaofeng~Chen,~\IEEEmembership{Member,~IEEE,} 
Jiaxing~Xu, Zongrui~Li, Yiping~Ke,~\IEEEmembership{Member,~IEEE,} 
Xudong~Jiang,~\IEEEmembership{Fellow,~IEEE,}
Yingchen~Yu, 
Yunqing~Zhao\textsuperscript{\Letter}\thanks{\Letter~\textit{Corresponding Author: yunqing.z.0817@gmail.com}.}, 
Song~Bai,~\IEEEmembership{Member,~IEEE}
\thanks{Mengcheng Lan, Jiaxing Xu, Yiping Ke are with College of Computing and Data Science, Nanyang Technological University. Zongrui Li, Xudong Jiang are with School of Electrical and Electronic Engineering, Nanyang Technological University. Chaofeng Chen is with School of Artificial Intelligence, Wuhan University. Yingchen Yu, Yunqing Zhao, Song Bai are with ByteDance.\!}
%
}

%
%

\markboth{IEEE Transactions,~2025}%
{Meng \MakeLowercase{\textit{et al.}}:}
%



\maketitle

\begin{abstract}
Multimodal Large Language Models (MLLMs) have shown exceptional capabilities in vision-language tasks. However, effectively integrating image segmentation into these models remains a significant challenge. 
In this work, we propose a novel \textit{text-as-mask} paradigm that casts image segmentation as a text generation problem, eliminating the need for additional decoders and significantly simplifying the segmentation process.
Our key innovation is \textit{\textmask}, a new textual representation of segmentation masks where each image patch is mapped to its corresponding text label.
We first introduce \emph{image-wise \textmask}, a patch-aligned textual representation of segmentation masks that integrates naturally into the language modeling pipeline. To enhance efficiency, we introduce the Row-wise Run-Length Encoding (\rle), which compresses redundant text sequences, reducing the length of \textmask by 74\% and accelerating inference by $3\times$, without compromising performance. Building upon this, our initial framework \name achieves strong segmentation performance across a wide range of vision tasks.
To further improve granularity and compactness, we propose \emph{box-wise \textmask}, which localizes regions of interest using bounding boxes and represents region masks via structured mask tokens called semantic bricks. This leads to our refined model, \namepp, which formulates segmentation as a next-brick prediction task, combining precision, scalability, and generative efficiency.
Comprehensive experiments on natural and remote sensing datasets show that \namepp consistently outperforms state-of-the-art models across diverse benchmarks without any task-specific fine-tuning, while remaining 
compatible with existing MLLM backbones. 
Our work highlights the effectiveness, scalability, and generalizability of text-driven image segmentation within the MLLM framework.

\end{abstract}


%
\IEEEpeerreviewmaketitle

\section{Introduction}\label{sec:introduction}
\IEEEPARstart{M}{ultimodal} Large Language Models (MLLMs)~\cite{yin2024survey} have significantly extended the capabilities of powerful Large Language Models (LLMs) into the visual domain. 
Recent advancements demonstrate their remarkable ability to perform natural language-based interaction and reasoning over visual inputs~\cite{liu2024visual, lu2024deepseek, liu2024improved, Qwen-VL, chen2024far}.
As a result, MLLMs are increasingly applied to a wide spectrum of vision-centric tasks, including image generation~\cite{song2024moma, wang2024genartist}, object detection~\cite{wang2024visionllm,ma2024groma,wu2024towards,zhang2023next} and semantic segmentation~\cite{li2024transformer,lan2023smooseg,lai2024lisa}.
Despite these advances, seamlessly integrating MLLMs with these tasks, particularly in dense prediction tasks like semantic segmentation, remains challenging due to the intrinsic differences between natural language and visual modalities.

\begin{figure*}[t]
    \newcommand{\colwid}{0.32\linewidth}
    \centering
    \includegraphics[width=0.85\linewidth]{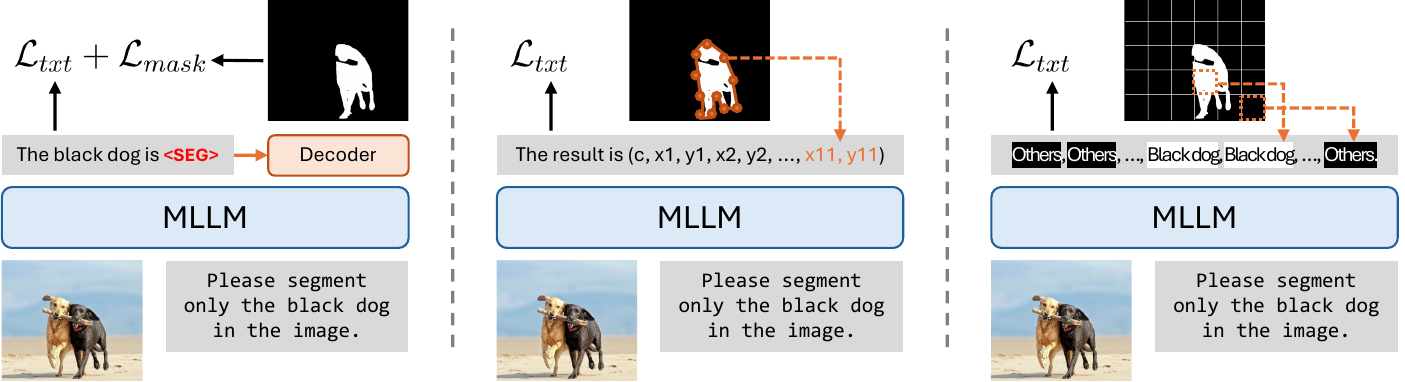}
    \makebox[\colwid]{\small (a) \textit{embeddings-as-mask}}
    \makebox[\colwid]{\small (b) \textit{polygon coordinates}}
    \makebox[\colwid]{\small (c) \textit{text-as-mask (ours)}}
    \captionsetup{font={footnotesize}}
    \caption{Different paradigms of MLLMs based image segmentation: (a) \textit{embeddings-as-mask} paradigm that relies on additional segmentation decoder and loss (\textit{e.g.}, LISA \cite{lai2024lisa}); (b) \textit{polygon coordinates} for image segmentation (\textit{e.g.}, VisionLLM \cite{wang2024visionllm} and VistaLLM \cite{pramanick2024jack}); (c) our \textit{text-as-mask} paradigm that relies on semantically consistent text sequences.}
    \label{fig:teaser}
    \vspace{-4mm}
\end{figure*}

A prevalent solution adopted by recent studies~\cite{lai2024lisa, xia2024gsva, zhang2024groundhog, he2024multi, ren2024pixellm, rasheed2024glamm, zhang2023next, wu2024towards} involves appending additional visual decoders (\eg, SAM \cite{kirillov2023segment}) on top of MLLMs, as illustrated in \Cref{fig:teaser}\,(a). 
In such framework, an MLLM primarily functions as a multimodal encoder that interprets user queries containing implicit or explicit references to regions of interest in an image.
It then generates a special \texttt{$<$seg$>$} token that serves as a semantic cue and is processed jointly with visual features by a mask decoder to yield the final segmentation mask.
While this paradigm has proven effective, it presents several drawbacks:
1) it complicates the end-to-end training pipeline with additional loss functions; 
2) it requires careful modifications to MLLM architectures, leading to unexpected challenges when scaling up the training; 
and 3) it remains fundamentally discriminative, underutilizing the inherent generative capabilities of LLMs.

Another line of work~\cite{wang2024visionllm, xiao2024florence, pramanick2024jack} represents segmentation masks as polygon coordinate sequences that can be decoded in an autoregressive manner, aligning more closely with the language modeling paradigm.
Notable examples include VisionLLM~\cite{wang2024visionllm} and VistaLLM~\cite{pramanick2024jack}, illustrated in \Cref{fig:teaser}\,(b).
Despite their conceptual elegance, these models often exhibit degraded performance, as LLMs struggle to associate coordinate sequences with accurate spatial shapes. 
This challenge has prompted the reintroduction of task-specific segmentation modules in improved variants like VisionLLMv2~\cite{wu2024visionllm}.
These limitations underscore the pressing need for more effective strategies to unlock the full potential of MLLMs for segmentation tasks.
Such method should adhere to the next-token prediction paradigm of MLLMs for easier optimization, minimize architectural modifications for scalability, and fully leverage text generation capabilities of LLMs.

In this paper, we introduce a novel \textit{text-as-mask} paradigm that casts image segmentation as a text generation problem, which significantly simplifies the segmentation process, as illustrated in \Cref{fig:teaser}\,(c).
At the core of this paradigm is a novel sequence representation of segmentation masks. 
Instead of using index masks or numerical coordinates, we map each flattened patch of the image to its corresponding text description (\textit{e.g.}, a semantic label, a short phrase, or a long sentence), forming a purely textual representation of images, named as \textbf{image-wise \textmask} (I-SD).
This representation offers several advantages: 
1) a unified sequence representation seamlessly integrated into the auto-regressive training pipeline of MLLMs, making joint optimization with text tasks easier; 
2) no architectural changes are required, allowing full utilization of existing MLLMs training infrastructure, making it ideal for scaling up; 
3) support for large label vocabularies, equivalent to semantic words; 
and 4) flexible switching between different kinds of image segmentation tasks.

Building upon our text-as-mask paradigm and image-wise \textmask, we present our conference paper \textbf{\name}~\cite{lan2024text4seg}, a decoder-free segmentation framework that fully leverages the generative capabilities of MLLMs. 
Inspired by ViT~\cite{vit}, we demonstrate that \textit{representing an image with 16 $\times$ 16 semantic words, i.e., $256$ length of \textmask, is sufficient to achieve satisfactory results}.
To improve efficiency, we introduce the Row-wise Run-Length Encoding (\rle), which losslessly compresses the repeated descriptors within each image row while preserving the spatial structure. 
Without compromising performance, \rle achieves a 74\% reduction in \textmask\ length and speeds up inference by $3\times$ on average.
To further enhance performance, we optionally apply an off-the-shelf mask refiner, \eg, SAM, as a post-processing step to obtain pixel-level segmentation masks.

While the image-wise \textmask offers a global, patch-aligned representation well-suited for dense semantic segmentation, it still exhibits certain limitations:
1) repetitive textual descriptions, especially long sentences, inflate the sequence length and limit resolution scalability; 
2) background tokens (\eg, \texttt{"others"}) dominate the \textmask, especially when segmenting small foreground objects in large scenes; 
and 3) it relies on explicit text labels for each image patch, making it less effective for reasoning-driven segmentation tasks without predefined semantics.

To address this, we propose a more focused and compact formulation: \textbf{box-wise \textmask} (B-SD).
This approach first localizes regions of interest using \textit{tagged bounding boxes}, then it generates segmentation masks within each region using the semantic descriptors.
By explicitly coupling \textit{where} to look and \textit{what} to segment, 
B-SD encapsulates both spatial and semantic information within a unified, autoregressive text format.
B-SD eliminates the receptive textual descriptions by a single label tag, and minimizes the overhead of dense background tokens by the bounding box constraint.
To further enhance expressiveness and efficiency, we extend the MLLM vocabulary with structured mask tokens, which we refer to as \textit{semantic bricks}, (\eg, \{\texttt{fg1}, \texttt{fg2}, $\ldots$, \texttt{fg63}\}, \{\texttt{bg1}, \texttt{bg2}, $\ldots$, \texttt{bg63}\}), enabling segmentation to be interpreted as symbolic plotting on a $64 \times 64$ canvas.
Ultimately, this leads to our \textit{next-brick prediction} framework: \textbf{\namepp}, which generates box-wise \textmask sequences brick-by-brick.
Surprisingly, \namepp with $64 \times 64$ B-SD not only achieves significantly finer-grained segmentation compared to $16 \times 16$ I-SD in \name, but also reduces the overall sequence length.
Additionally, \namepp preserves the elegance of generative language modeling with improved precision and scalability, opening new possibilities for dense prediction via pure text generation.

With the proposed semantic descriptors, training MLLMs for segmentation requires minimal additional effort. 
We begin by constructing instruction-following data from existing segmentation datasets, transforming the vanilla semantic masks into the \textmask format, and then fine-tuning the model using query-response conversations. 
In our initial work~\cite{lan2024text4seg}, \name was fine-tuned and evaluated individually on each downstream segmentation task, demonstrating strong performance without any architectural changes.

In this work, we take a significant step further with \namepp by embracing a unified and generalizable framework, tailored for \textit{conversational} image segmentation. 
We curate a large-scale, diverse training corpus that integrates a wide range of visual tasks, including referring expression segmentation, generalized referring expression segmentation, single- and multi-object reasoning segmentation, as well as visual grounding and understanding, across both natural and remote sensing imagery domains. 
This setup enables us to train a single, versatile MLLM segmentation model capable of performing robustly across tasks and domains \textit{without the need for task-specific fine-tuning}. 
Our experiments demonstrate that \namepp can seamlessly integrate segmentation capabilities into existing MLLM architectures, such as Qwen2-VL \cite{wang2024qwen2}, Deepseek-VL2 \cite{wu2024deepseek}, and InternVL3 \cite{zhu2025internvl3}, \textit{without any architectural modifications}.
Without bells and whistles, \namepp consistently achieves superior performance to previous state-of-the-art methods, highlighting its efficiency, flexibility, and robustness.
In summary, our key contributions are as follows:
\begin{itemize}
    \item[] \hspace*{-\leftmargin} \textbf{In \name}:
    \vspace{3pt}
    \item  We propose a novel \textit{text-as-mask} paradigm formulating image segmentation as a text generation problem, which fully leverages the text generation capabilities of MLLMs.
    \item We introduce image-wise \textmask, a patch-aligned textual representation of segmentation masks, and an efficient Row-wise Run-Length Encoding (\rle) to reduce sequence length and speed up inference. Built on this, \name achieves strong performance across diverse image segmentation tasks when conducting task-specific fine-tuning on different benchmarks.
    \vspace{3pt}
    \item[] \hspace*{-\leftmargin} \textbf{In \namepp}:
    \vspace{3pt}
    \item We further propose box-wise \textmask, a focused region-level representation that unifies visual grounding and segmentation by jointly leveraging bounding boxes and semantic descriptors.
    This innovation is further bolstered by the introduction of semantic bricks, which significantly improve the compactness and decoding efficiency, enabling the generation of finer-grained, scalable segmentation masks with exceptional precision.
    \item We develop \namepp, a unified and generalizable framework built upon our next-brick prediction. \namepp integrates diverse image segmentation and understanding tasks, allowing the training of a single, versatile model without any task-specific fine-tuning.
    \item Our method achieves state-of-the-art performance and robustness across a wide range of vision-centric tasks. Additionally, it demonstrates strong compatibility with various Multimodal Large Language Model (MLLM) backbones, highlighting its versatility and extensibility.

\end{itemize}

\section{Related Work}\label{sec:relatedwork}
\subsection{Multimodal Large Language Models}
MLLMs are typically developed by enhancing large language models (LLMs) with visual perception modules, which can generate coherent textual conversations grounded in multimodal inputs.
For instance, Flamingo~\cite{alayrac2022flamingo} introduces the Perceiver Resampler, which connects a pre-trained vision encoder with LLMs for effective few-shot learning~\cite{awadalla2023openflamingo, li2025otter}. 
BLIP-2~\cite{li2023blip} and InstructBLIP~\cite{dai2023instructblip} bridge the modality gap using a lightweight Querying Transformer (Q-Former), demonstrating enhanced performance on zero-shot vision-to-language tasks.
The LLaVA series~\cite{liu2024visual,liu2024improved} employs a linear layer or MLP as a modality connector, trained on multimodal language-image instruction-following data generated with GPT-4, showcasing notable capabilities in multimodal chat interactions. 
In contrast, Qwen-VL~\cite{Qwen-VL} and mPLUG-Owl2~\cite{ye2024mplug} explore feature compression to a fixed length through cross-attention mechanisms with learnable queries.
Recent advancements in multimodal modeling~\cite{liu2024llava, guo2024llava, li2025llavaonevision, li2024mini, li2024monkey, lin2024sphinx, wei2024instructseg, chen2024expanding, zhu2025internvl3, wang2024qwen2, bai2025qwen2, wu2024deepseek} have focused on enhancing visual encoding through high-resolution inputs. 
For example, 
LLaVA-NEXT~\cite{liu2024llava} and LLaVA-OneVision~\cite{li2025llavaonevision} utilize the AnyRes scheme to accommodate high-resolution image inputs.
In contrast, Qwen2-VL~\cite{wang2024qwen2} and Qwen2.5-VL~\cite{bai2025qwen2} support native dynamic resolution through the introduction of a 2D-RoPE mechanism.
Despite their strong visual capabilities in tasks such as general visual question answering, document understanding, OCR, and visually grounded agent applications, these MLLMs remain limited in their ability to perform dense prediction tasks, such as image segmentation.
In this work, we present \name and \namepp to endow existing MLLMs with image segmentation capabilities based on instruction tuning, \textit{without necessitating any changes to their architecture}.

\subsection{MLLMs for Visual Segmentation} 
\noindent \textbf{Discriminative Models.}
Recent advancements have enabled MLLMs to support image segmentation by incorporating task-specific modules \eg, additional image encoder or mask decoder \cite{kirillov2023segment}.
In these frameworks, MLLM primarily interprets user queries, which may contain implicit or explicit references to target objects in the image, and generate a special \texttt{$<$seg$>$} token that serves as a text-based cue.
This cue, along with the visual features extracted by the image encoder, is then passed to the mask decoder, which produces the corresponding binary segmentation mask.
A line of work, 
including LISA~\cite{lai2024lisa} and its successors~\cite{xia2024gsva, rasheed2024glamm, zhang2024omg, ren2024pixellm,zhang2023next, he2024multi, jang2025mmr, wang2025segllm} adopt this \textit{embedding-as-mask} paradigm and demonstrates strong performance on tasks such as reasoning segmentation and referring expression segmentation.
This multimodal segmentation paradigm has also been extended to the remote sensing domain \cite{li2025segearth, ou2025geopix}.
In contrast, UFO~\cite{tang2025ufo} and Pixel-SAIL~\cite{zhang2025pixel} propose to directly extract image embeddings from MLLMs and generate segmentation masks by computing the similarity between the image embeddings and text (\ie, \texttt{$<$seg$>$}) embedding.
However, this discriminative segmentation paradigm often complicates the end-to-end training pipeline due to the need for additional loss functions and architectural components.

\noindent \textbf{Generative Models.}
Another major line of research investigates the strong generative capabilities of MLLMs for segmentation tasks.
For example, HiMTok~\cite{wang2025himtok} and ALTo~\cite{wang2025alto} use MLLMs to generate discrete mask tokens, which are then decoded via a mask detokenizer to produce fine-grained segmentation masks, achieving decent performance on various segmentation tasks.
Other approaches~\cite{wang2024visionllm, xiao2024florence, pramanick2024jack} directly predict the polygon coordinates delineating object boundaries.
However, these generative approaches often suffer from limited performance, as MLLMs may struggle to associate geometric representations (\eg, polygon coordinates) with precise object shapes, leading to inaccurate or coarse masks.


\subsection{MLLMs for Visual Grounding}
Visual grounding aims to localize specific objects in an image based on a natural language instructions, serving as a fundamental task that bridges vision and language modalities.
Traditional approaches~\cite{chen2022pixseq, liu2024grounding, yan2023universal} typically frame it as a detection problem, employing task-specific architectures that combine object detection with language encoders to align textual phrases with corresponding image regions.
Recent advances in MLLMs have enabled more flexible and general-purpose solutions to visual grounding.
For instance, Kosmos-2~\cite{peng2024grounding} and Shikra~\cite{chen2023shikra} discretize spatial locations by quantizing bounding boxes into either location tokens or numeric position representations.
More recently, generalist MLLMs~\cite{wu2024deepseek, zhu2025internvl3, ma2024groma, fei2024vitron,bai2025qwen2} have demonstrated the ability to directly predict bounding boxes as outputs in response to textual queries.
In this work, we distinguish our approach from prior studies by binding bounding boxes with semantic masks, which provide denser and more informative supervision signals. This enables the model to learn visual-linguistic alignments with improved granularity, ultimately enhancing both grounding precision and segmentation quality.

\section{Methodology}\label{sec:method}

\begin{figure}[t]
\centering
\includegraphics[width=0.3\textwidth]{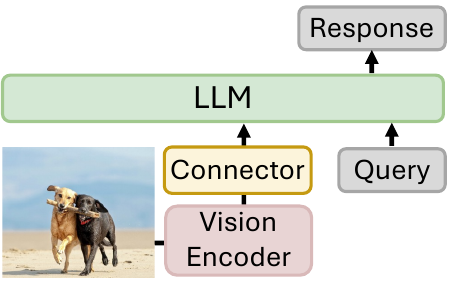}
\captionsetup{font={footnotesize}}
\caption{
An illustration of a popular MLLM architecture.
}
\label{fig:mllm}
\end{figure}

\begin{figure*}[t]
    \centering
    \includegraphics[width=1.0\linewidth]{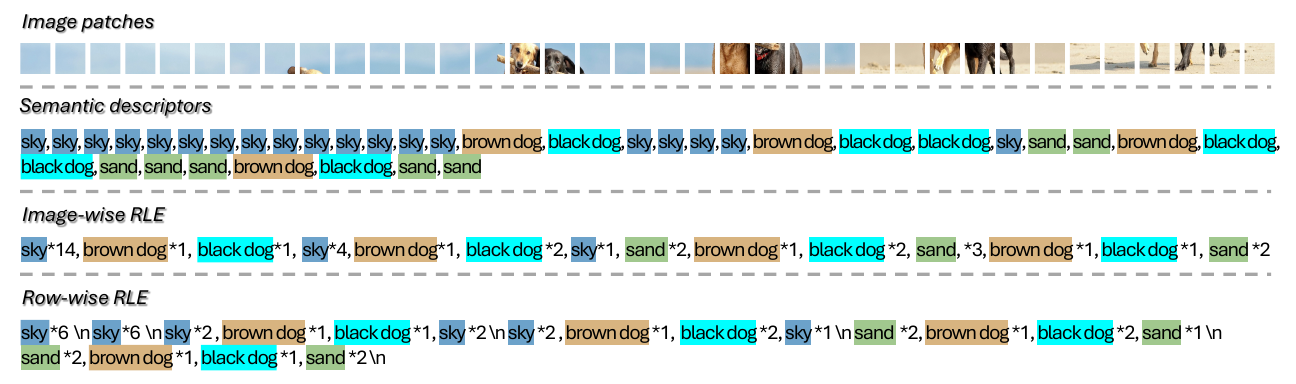}
    \captionsetup{font={footnotesize}}
    \caption{An illustration of image patches, image-wise \textmask and two token compression techniques.}
    \label{fig:descriptors}
\end{figure*}

\subsection{Preliminary}
\label{preliminary}
Multimodal Large Language Models (MLLMs)~\cite{yin2024survey} refer to the LLM-based models with the ability to process, reason, and generate response from multimodal input information.
Typically, as shown in \Cref{fig:mllm}, 
an MLLM can be abstracted into three main components:
\begin{itemize}[leftmargin=*]
\item \textit{Vision Encoder:} A pre-trained vision encoder (\eg, CLIP~\cite{radford2021learning} or SigLIP~\cite{zhai2023sigmoid}) that transforms input images into a sequence of visual tokens.
\item \textit{Language Model:} A pre-trained large language model (LLM), such as Qwen3~\cite{yang2025qwen3} or LLaMA~\cite{touvron2023llama}, responsible for understanding and generating natural language outputs through next-token prediction.
\item \textit{Modality Connector:} A bridging module that aligns visual and textual modalities. This is often implemented using lightweight architectures like a two-layer MLP~\cite{liu2024improved} or cross-attention mechanisms~\cite{Qwen-VL}, enabling effective fusion of visual features into the LLM’s context window.
\end{itemize}

\subsection{\name with Image-wise Semantic Descriptors}
\label{descirptors}
\subsubsection{Definition of image-wise \textmask}
Inspired by the patch-based representation of Vision Transformers (ViT)~\cite{vit},
our \textmask encode segmentation masks into a sequence of semantic tokens spatially aligned with visual patches.
As illustrated in \Cref{fig:descriptors},
the process begins by splitting the image into a grid of fixed-size patches (\eg, $16 \times 16$) and flattening them, resulting in 256 non-overlapping regions. 
Each patch is then represented by its corresponding semantic descriptor. 
A descriptor can be as simple as a semantic label (\eg, ``sky", ``sand"), a phrase (\eg, ``brown dog", ``black dog"), or even a more complex textual description (\eg, ``a brown dog on the left side") for intricate scenes. 
This design choice transforms an image into a sequence of image-wise \textmask (I-SD) of length $256$,
which meets the requirements for integrating image segmentation into MLLMs and offers several key advantages:
\begin{itemize}[leftmargin=*]
\item It naturally aligns with the next-token prediction paradigm of existing LLMs, reframing image segmentation as a standard generative language modeling task, facilitating easier optimization.
\item It requires no modifications to existing MLLMs architecture, making it easily scalable and compatible with existing training infrastructures.
\item It adopts a \textit{text-as-mask} paradigm, 
fully using the text generation capabilities of LLMs for image segmentation.
\end{itemize}

One of the key limitations of full-length image-wise \textmask lies in its substantial token length, which stems from the spatial redundancy inherent in pixel-aligned representations.
For instance, on the RefCOCO~\cite{kazemzadeh2014referitgame} dataset, the average token length of $256$-I-SD is $583$, requiring approximately 19s on an NVIDIA V100 GPU for a single round of referring expression segmentation.

\subsubsection{Image-wise RLE} To address this inefficiency, we explore the simple Run-Length Encoding (RLE) \cite{golomb1966run},
a classic compression technique, to reduce redundant tokens in the sequence.
A naïve solution is to apply RLE directly across the entire \textmask sequence, referred to as Image-wise RLE (I-RLE), as shown in \Cref{fig:descriptors}.
However, we empirically found that it results in a notable performance degradation, dropping nearly 4 cIoU on the RefCOCO \texttt{val} split.
This indicates that compressing the semantic descriptors globally may disrupt essential two-dimensional spatial patterns that MLLMs rely on for accurate segmentation.

\subsubsection{Row-wise RLE} To mitigate this issue, we propose a novel Row-wise Run-Length Encoding (\rle) technique. 
As shown in \Cref{fig:descriptors}, R-RLE compresses adjacent repeated descriptors within each row of the semantic descriptor grid, with each row separated by ``\texttt{$\setminus$n}''. 
This approach reduces the token length from 583 to 154 on average while preserving more spatial information.
Importantly, \rle demonstrates no performance degradation compared to the full-length \textmask, and significantly enhances the inference speed.

\subsubsection{Visual Instruction Tuning}
\label{data}
\begin{figure*}[t]
\begin{minipage}{0.7\textwidth}
\centering
\includegraphics[width=1.0\linewidth]{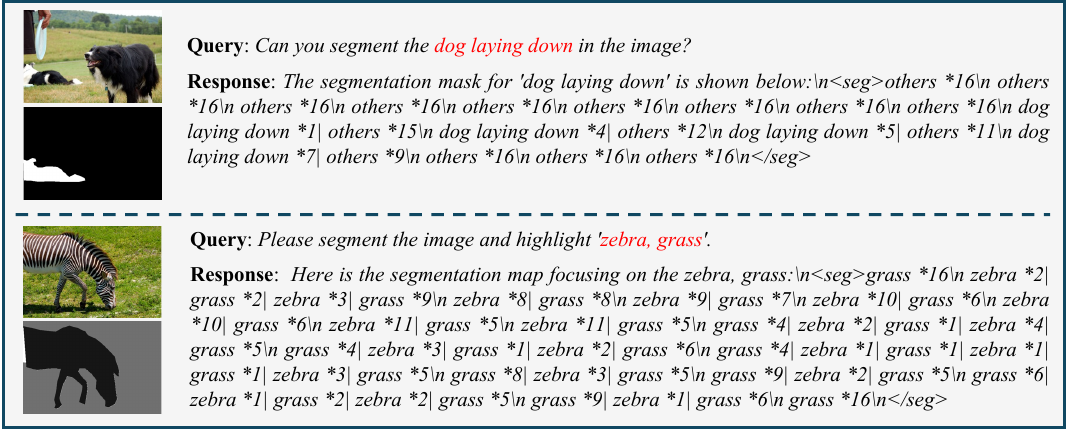}
\captionsetup{font={footnotesize}}
\caption{Visual instruction tuning data based on image-wise semantic descriptors.}
\label{fig:data}
\end{minipage}
\hfill
\begin{minipage}{0.27\textwidth}
\centering
\includegraphics[width=1.0\linewidth]{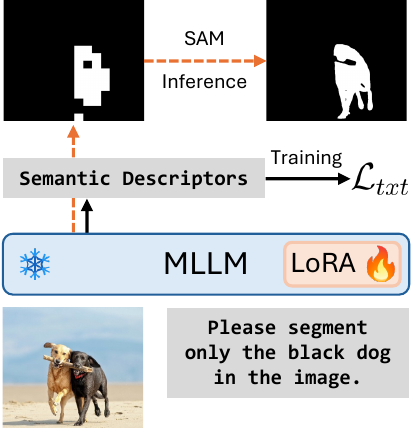}
\captionsetup{font={footnotesize}}
\caption{\name architecture.}
\label{fig:framework}
\end{minipage}
\end{figure*}

Building upon the proposed I-SD, we construct visual instruction data by repurposing existing segmentation annotations into an instruction-following format. 
\Cref{fig:data} illustrates a few examples from referring expression segmentation.
Given an $<$image, mask$>$ pair, 
we first resize the segmentation mask to a fixed $16 \times 16$ resolution and flatten it into a 1D sequence.
The numerical labels in the sequence are then replaced with their corresponding text labels to create full-length semantic descriptors.
To reduce redundancy and improve efficiency, we further apply R-RLE to compress the sequence, with descriptors separated by ``\texttt{$|$}'' and rows separated by ``\texttt{$\setminus$n}''.
Finally, we wrap the image input, textual labels (e.g., class names or referring expressions), and the compressed descriptors into an instruction-following query-response format as shown below:
\begin{tcolorbox}[colback=gray!5!white, colframe=gray!65!black, title=Instruction-response template, sharp corners=south]
\textbf{Query:} \texttt{\textless image\textgreater} Can you segment the \texttt{\textless text labels\textgreater} in the image?

\vspace{0.5em}
\textbf{Response:} The result is: \texttt{\textbackslash n \textless seg\textgreater image-wise semantic descriptors\textless/seg\textgreater.}
\end{tcolorbox}

\noindent
Here, \texttt{$<$image$>$} is a placeholder for the visual input tokens, while \texttt{$<$seg$>$} and \texttt{$</$seg$>$} are special markers indicating the start and end of the \textmask sequence.

\subsubsection{Text4Seg Architecture}
With such text-only response design, \name can be seamlessly integrated with existing MLLMs without any architectural modifications, as shown in \Cref{fig:framework}.
We use Low-Rank Adaptation (LoRA)~\cite{hu2022lora} to fine-tune the MLLMs on our visual instruction data, using its original auto-regressive training objective $\mathcal{L}_{txt}$~\cite{liu2024visual}. 
In contrast to existing models~\cite{lai2024lisa,zhang2024groundhog,rasheed2024glamm}, which typically rely on Continued Pre-Training (CPT) with large, mixed datasets to fuse the architectures before fine-tuning on specific downstream tasks, we directly apply Supervised Fine-Tuning (SFT) on the downstream tasks.
During inference, to obtain a better pixel-level semantic mask, we optionally apply SAM as the mask refiner with our generated coarse mask as its prompt.

\subsection{\namepp with Box-wise Semantic Descriptors}
\label{box-wise-semantic-descriptors}
\subsubsection{Motivation}
While the image-wise \textmask provide a global, patch-aligned textual representation well-suited for dense semantic segmentation, and \name demonstrates strong performance across individual segmentation tasks, it still faces several limitations:
\begin{itemize}[leftmargin=*]
\item \textit{Redundant textual descriptions:} Repetitive semantic descriptors, especially full-sentence annotations, result in unnecessarily long token sequences. This increases computational overhead and hinders scalability to higher-resolution inputs, ultimately limiting fine-grained segmentation performance.

\item \textit{Background token dominance:} In scenes with large backgrounds and small foreground objects, the \textmask are often dominated by background tokens (\eg, \texttt{"others"}), reducing the semantic density and expressiveness of the representation.

\item \textit{Dependency on explicit labels:} Image-wise \textmask require explicit semantic labels for each patch, which becomes limited in reasoning-driven tasks (\eg, reasoning segmentation) where semantics are context-dependent or inferred. This restricts generalization to tasks beyond fixed label spaces.
\end{itemize}

To address these limitations, we introduce a more compact representation: box-wise \textmask (B-SD).

\begin{figure*}[h]
    \centering
    \includegraphics[width=0.85\linewidth]{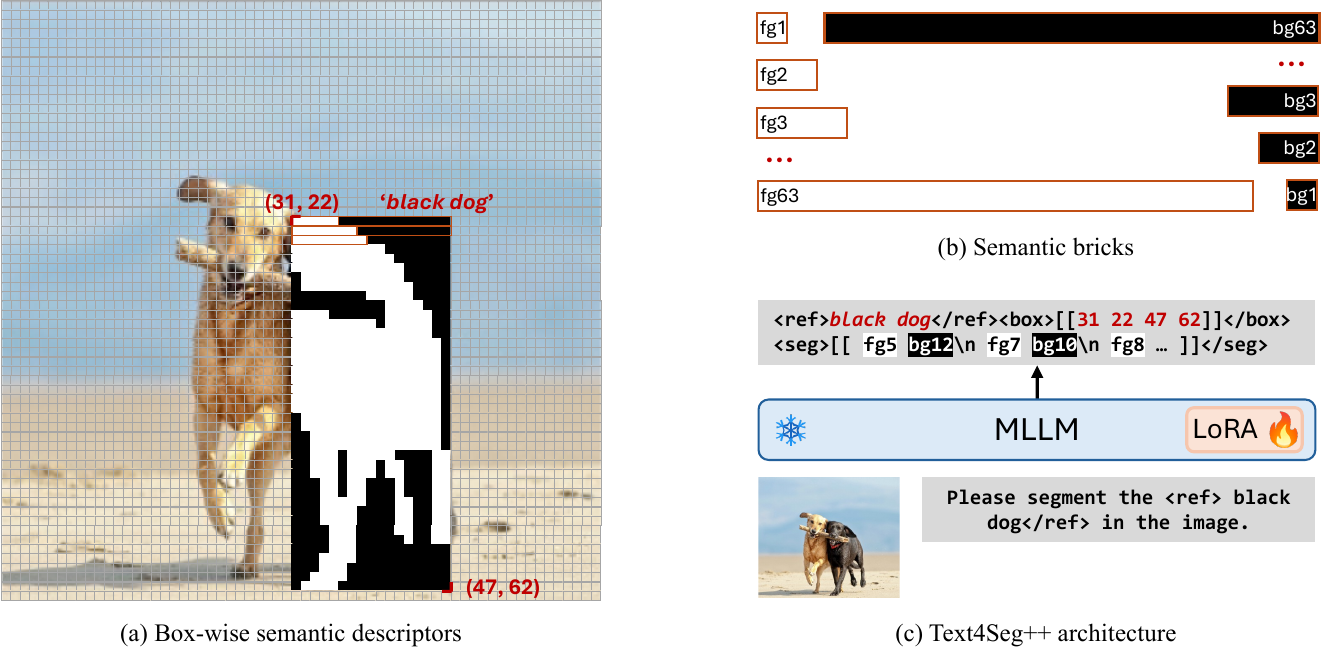}
    \captionsetup{font={footnotesize}}
    \caption{An illustration of (a) box-wise \textmask for images, (b) semantic bricks and (c) \namepp framework.}
    \label{fig:text4seg++}
\end{figure*}

\begin{figure}[t]
\centering
\includegraphics[width=0.35\textwidth]{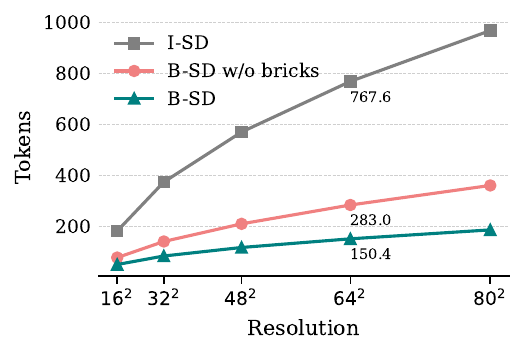}
\captionsetup{font={footnotesize}}
\caption{Token counts with varying resolution of semantic descriptors for three configurations: I-SD, B-SD, and B-SD without semantic bricks.}
\label{fig:tokens_resolution}
\end{figure}

\subsubsection{Definition of box-wise \textmask} 
We reformulate image segmentation as a two-step process: visual grounding followed by visual segmentation. 
Based on this formulation, we introduce a novel representation, box-wise semantic descriptors, to describe each segmented instance in a compact and structured textual format, as illustrated in \Cref{fig:text4seg++}\,(a).
Specifically, each instance is represented using the following syntax:
\begin{tcolorbox}[colback=gray!5!white, colframe=gray!65!black, title=Box-wise semantic descriptor, sharp corners=south]
\footnotesize
\texttt{\textless ref\textgreater text labels\textless /ref\textgreater \textless box\textgreater [[x1 y1 x2 y2]] \textless /box\textgreater \textless seg\textgreater semantic descriptors\textless/seg\textgreater}
\end{tcolorbox}
\noindent
where
\texttt{\textless ref\textgreater}, \texttt{\textless /ref\textgreater}, \texttt{\textless box\textgreater}, \texttt{\textless /box\textgreater}, \texttt{\textless seg\textgreater} and \texttt{\textless /seg\textgreater} are special tokens.
This representation consists of three core components:
\begin{itemize}[leftmargin=*]
\item \texttt{<ref>...</ref>}: A natural language referring the expression or category label that provides semantic grounding for the object (\eg, “black dog”, “the person on the left”).
To support scenarios without explicit labels, such as reasoning segmentation where object identity is context-dependent, we introduce abstract region identifiers (\eg, \texttt{roi0}, \texttt{roi1}, ...). 
This strategy mitigates the issue of repetitive semantic descriptors and alleviates the reliance on explicit labels.

\item \texttt{<box>...</box>}: The bounding box coordinates of the object, formatted as \texttt{[[x1 y1 x2 y2]]}, where \texttt{(x1, y1)} and \texttt{(x2, y2)} represent the top-left and bottom-right corners, respectively.
Each coordinate is quantized into one of 64 discrete bins, aligning with the resolution of the semantic descriptors.
By localizing the region of interest, this structure effectively reduces the dominance of background tokens and ensures that the model focuses on semantically meaningful regions.

\item \texttt{<seg>...</seg>}: A concise semantic descriptors of the object mask.
\end{itemize}

\begin{figure}[t]
\centering
\includegraphics[width=0.49\textwidth]{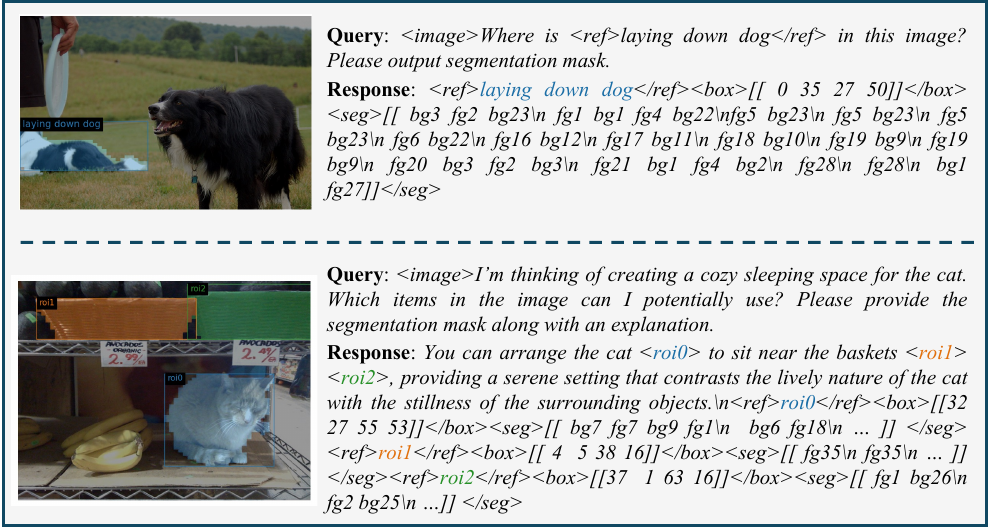}
\captionsetup{font={footnotesize}}
\caption{Visual instruction tuning data based on box-wise semantic descriptors.}
\label{fig:text4segpp_examples}
\end{figure}

Together, these components provide a compact yet informative textual representation that supports efficient and generalizable dense segmentation.

\subsubsection{Next Brick Prediction}
While our 
\rle significantly reduces the length of semantic descriptors, there is still room for further compression.
Empirically, we observe that each semantic block can span multiple tokens. 
For instance, the string \texttt{others*16} is tokenized into four separate tokens, ``\texttt{others}", ``\texttt{*}", ``\texttt{1}" and ``\texttt{6}", when using the Qwen \cite{yang2025qwen3} tokenizer. 
This token-level granularity introduces unnecessary overhead in both sequence length and decoding time.

To further enhance the compactness and decoding efficiency of the semantic descriptors, we introduce a set of special tokens, referred to as \textit{semantic bricks}, as illustrated in \Cref{fig:text4seg++}\,(b).
Specifically, we construct a vocabulary of 126 bricks: 63 white foreground bricks (denoted as \texttt{fg1}, \texttt{fg2}, ..., \texttt{fg63}) and 63 black background bricks (denoted as \texttt{bg1}, \texttt{bg2}, ..., \texttt{bg63}). 
Each brick corresponds to a binary segment of varying lengths, encoding the object mask in a compact and interpretable form.
To reconstruct the full mask from these bricks,
we adopt a sequential generation strategy, referred to as \textit{next brick prediction}, where the model predicts one brick at a time to construct the binary mask. 
The bricks are arranged from left to right and top to bottom, mimicking the raster-scan order of a 2D mask. 
This design not only reduces token count and improves inference speed but also aligns well with the autoregressive generation paradigm of large language models.

We quantitatively compare the token lengths of image-wise and box-wise \textmask representations using the Qwen tokenizer on the RefCOCO dataset, as shown in \Cref{fig:tokens_resolution}. 
The results reveal that the box-wise formulation is significantly more compact. 
For instance, at a resolution of $64 \times 64$, the average token length of B-SD without semantic bricks is 283.0, substantially shorter than the 767.6 tokens required by the image-wise counterpart. 
Furthermore, incorporating semantic bricks into the B-SD reduces the token length even further to 150.4. 
These findings demonstrate the token efficiency and scalability of our box-wise \textmask.

In \Cref{fig:text4segpp_examples}, we present two qualitative examples of visual instruction response. 
These examples illustrate that our proposed B-SD is effective not only for image segmentation tasks with explicit labels (\eg, referring expression segmentation) but also for tasks without explicit supervision, such as reasoning-driven segmentation. 
This highlights the generality and versatility of our framework in both low-level and high-level segmentation scenarios.

\subsubsection{\namepp Architecture}
As illustrated in \Cref{fig:text4seg++}\,(c), \namepp maintains the architectural simplicity of \name by leveraging a pure text-based output format, enabling seamless integration with existing MLLMs without any structural modifications. 
Specifically, \namepp generates high-resolution B-SD as autoregressive text responses, enabling more fine-grained and spatially precise segmentation.
To support this higher-resolution output, it is essential that the input image also maintains sufficient resolution. 
Consequently, we adopt MLLM architectures capable of processing high-resolution visual inputs, such as Qwen2-VL-7B~\cite{wang2024qwen2} and InternVL3-8B~\cite{zhu2025internvl3}, ensuring high fidelity in both visual encoding and textual decoding.
We use LoRA to perform post-training on the MLLMs using the pure generative language modeling loss.

\section{Experiments} \label{sec:experiments}
\subsection{Experiment Setup}

\subsubsection{Implementation Details}
Our methods are designed to be seamlessly integrated into existing Multimodal Large Language Model (MLLM) architectures.
For \name, we adopt InternVL2-8B~\cite{chen2024far} and LLaVA-1.5-13B~\cite{liu2024improved} as base models.
For \namepp, we employ Qwen2-VL-7B~\cite{wang2024qwen2}, which natively supports dynamic input resolutions and demonstrates higher optimization efficiency.
All MLLM architectures are kept unchanged during the experiments.
Additionally, we optionally incorporate SAMRefiner \cite{lin2025samrefiner} with a ViT-H backbone as an off-the-shelf mask refinement module.

Our methods are implemented using SWIFT framework~\cite{zhao2024swiftascalablelightweightinfrastructure}. 
\namepp is trained on 8 NVIDIA H100 GPUs with a global batch size of 128. 
We use the AdamW optimizer \cite{loshchilov2018decoupled}, starting with an initial learning rate of 2e-4, which follows a linear decay schedule after a warm-up phase with a ratio of 0.03. 
The weight decay is set to 0, and gradient norms are clipped at 1.0.
To minimize GPU memory usage, we fine-tune all models using LoRA with a rank of 128, along with ZeRO-1 stage memory optimization~\cite{rajbhandari2020zero}. 

\subsubsection{Evaluation Protocol}
We adopt a comprehensive set of metrics to evaluate the effectiveness of our proposed methods:
\begin{itemize}[leftmargin=*]
\item \textbf{cIoU} (Cumulative IoU)~\cite{wu2020phrasecut}: Computes the cumulative intersection over cumulative union across all samples. This metric favors larger objects by aggregating pixel-level overlap globally.
\item \textbf{gIoU} (Generalized IoU)~\cite{liu2023gres}: Calculates the mean per-image IoU across all samples. For no-target cases, true positive predictions are assigned an IoU of 1, while false negatives are assigned 0. This metric provides a balanced assessment of performance on both small and large objects.
\item \textbf{mIoU} (Mean IoU): Similar to gIoU, it computes the average per-image IoU across the dataset. 
\item \textbf{ACC@0.5}: Accuracy of the IoU between the predicted and ground truth bounding boxes thresholds at 0.5, reflecting the referring expression comprehension.
\item \textbf{Accuracy}: Measures the correctness of the model's answers in the visual question answering task, assessing multimodal understanding and reasoning capabilities.
\end{itemize}
This diverse set of metrics allows for a thorough evaluation of visual segmentation grounding and understanding.

\begin{table}[t]
\caption{\footnotesize\\ Statistics of the training data.
}
\label{tab:statistics}
\begin{center}
\tabcolsep3pt
\vspace{-3mm}
\begin{tabular}{lccc}
\toprule
\textbf{Dataset} & \textbf{Task Description} & \textbf{\#Image} & \textbf{\#Sample} \\
\midrule
COCO & panotic segmentation & 118k & 236k \\
refCOCO & \makecell{referring expression segmentation \\ referring expression comprehension} & 74k & 439k \\
grefCOCO & \makecell{generalized referring \\ expression segmentation} & 17k & 209k \\
Pix2Cap & referring expression segmentation & 18k & 152k \\
ReasonSeg & single-object reasoning segmentation & 209 & 1045 \\
MUSE & multi-object reasoning segmentation & 102k & 230k  \\
RRSIS\_D & \makecell{remote sensing referring \\ image segmentation \& grounding} & 12k & 61k \\
Earthreason & geospatial reasoning segmentation & 2.4k & 71k  \\
LLaVA-665k & visual understanding & 665k & 665k \\
\bottomrule
\end{tabular}
\vspace{-5mm}
\end{center}
\end{table}

\begin{table*}[t]
\vspace{-4mm}
\caption{\footnotesize\\ \textbf{Referring Expression Segmentation} results (\textbf{cIoU}) on \textbf{refCOCO (+/g)} \cite{kazemzadeh2014referitgame,mao2016generation} benchmarks. 
\textbf{Mask Dec.}: Mask decoder.
U: The UMD partition. 
FT: Models are finetuned on the joint training split of the referring expression segmentation datasets.
$^\dagger$ Model based on the 32$\times$ 32 I-SD without the mask refiner. 
$^\ddagger$ Model based on the 64$\times$ 64 B-SD without the mask refiner. 
The best results are highlighted in \best{Best}, while the second-best results are marked with \second{Second}.
}
\label{tab:res}
\begin{center}
\renewcommand\arraystretch{0.98}
\tabcolsep7pt
\begin{tabular}{llcccccccccc}
\toprule
\multirow{2}*{\textbf{Method}} & \multirow{2}*{\textbf{LLM}} & \multirowcell{2}{\textbf{Mask} \\ \textbf{Dec.}} & \multicolumn{3}{c}{\textbf{refCOCO}} & \multicolumn{3}{c}{\textbf{refCOCO+}} & \multicolumn{2}{c}{\textbf{refCOCOg}} & \multirow{2}*{\textbf{Avg.}}\\
 & & & \textbf{val} & \textbf{testA} & \textbf{testB} & \textbf{val} & \textbf{testA} & \textbf{testB} & \textbf{val(U)} & \textbf{test(U)} & \\
\midrule
\multicolumn{12}{c}{\textit{\textbf{Specialised Segmentation Models}}} \\
ReLA \scriptsize{\textcolor{gray}{\textup{[CVPR23]}}} \cite{liu2023gres} & BERT & \ding{51} &  73.8 & 76.5 & 70.2 & 66.0 & 71.0 & 57.7 & 65.0 & 66.0 & 68.3 \\
PolyFormer-L \scriptsize{\textcolor{gray}{\textup{[CVPR23]}}} \cite{liu2023polyformer} & BERT & \ding{55} & 76.0 & 78.3 & 73.3 & 69.3 & 74.6 & 61.9 & 69.2 & 70.2 & 71.6 \\
UNINEXT-L \scriptsize{\textcolor{gray}{\textup{[CVPR24]}}}
\cite{yan2023universal} & BERT & \ding{51} & 80.3 & 82.6 & 77.8 & 70.0 & 74.9 & 62.6 & 73.4 & 73.7 & 74.4 \\
LAVT \scriptsize{\textcolor{gray}{\textup{[TPAMI24]}}} \cite{yang2024language} & BERT & \ding{51} & 79.2 & 80.7 & 75.4 & 71.7 & 75.6 & 64.3 & 72.1 & 74.6 & 74.2 \\
\midrule
\multicolumn{12}{c}{\textit{\textbf{Generalist Segmentation Models (7B)}}} \\
NEXT-Chat (FT) \scriptsize{\textcolor{gray}{\textup{[ICML24]}}} \cite{zhang2023next} & Vicuna-7B & \ding{51} & 74.7 & 78.9 & 69.5 & 65.1 & 71.9 & 56.7 & 67.0 & 67.0 & 68.9 \\
LISA (FT) \scriptsize{\textcolor{gray}{\textup{[CVPR24]}}} \cite{lai2024lisa} & Vicuna-7B & \ding{51} & 74.9 & 79.1 & 72.3 & 65.1 & 70.8 & 58.1 & 67.9 & 70.6 & 69.9 \\
GSVA (FT) \scriptsize{\textcolor{gray}{\textup{[CVPR24]}}} \cite{xia2024gsva} & Vicuna-7B & \ding{51} & 77.2 & 78.9 & 73.5 & 65.9 & 69.6 & 59.8 & 72.7 & 73.3 & 71.4 \\
PixelLM \scriptsize{\textcolor{gray}{\textup{[CVPR24]}}} \cite{ren2024pixellm} & Vicuna-7B & \ding{51} & 73.0 & 76.5 & 68.2 & 66.3 & 71.7 & 58.3 & 69.3 & 70.5 & 69.2 \\
AnyRef (FT) \scriptsize{\textcolor{gray}{\textup{[CVPR24]}}} \cite{he2024multi} & LLaMA2-7B & \ding{51} & 76.9 & 79.9 & 74.2 & 70.3 & 73.5 & 61.8 & 70.0 & 70.7 & 72.2 \\
Groundhog \scriptsize{\textcolor{gray}{\textup{[CVPR24]}}} \cite{zhang2024groundhog} & LLaMA2-7B & \ding{51} & 78.5 & 79.9 & 75.7 & 70.5 & 75.0 & 64.9 & 74.1 & 74.6 & 74.2 \\ 
GLaMM (FT) \scriptsize{\textcolor{gray}{\textup{[CVPR24]}}} \cite{rasheed2024glamm} & Vicuna-7B & \ding{51} & 79.5 & 83.2 & 76.9 & 72.6 & 78.7 & 64.6 & 74.2 & 74.9 & 75.6 \\
SAM4MLLM \scriptsize{\textcolor{gray}{\textup{[ECCV24]}}} \cite{chen2024sam4mllm} & Vicuna-7B & \ding{51} & 79.8 & 82.7 & 74.7 & \second{74.6} & 80.0 & 67.2 & 75.5 & 76.4 & 76.4 \\
OMG-LLaVA (FT) \scriptsize{\textcolor{gray}{\textup{[NeurIPS24]}}} \cite{zhang2024omg} & InterLM2-7B & \ding{51} & 78.0 & 80.3 & 74.1 & 69.1 & 73.1 & 63.0 & 72.9 & 72.9 & 72.9 \\
VITRON (FT) \scriptsize{\textcolor{gray}{\textup{[NeurIPS24]}}} \cite{fei2024vitron} & Vicuna-7B & \ding{51} & 75.5 & 79.5 & 72.2 & 66.7 & 72.5 & 58.0 & 67.9 & 68.9 & 70.2 \\
M$^2$SA \scriptsize{\textcolor{gray}{\textup{[ICLR25]}}} \cite{jang2025mmr} & Vicuna-7B & \ding{51} & 74.0 & 76.8 & 69.7 & 63.1 & 67.2 & 56.1 & 67.0 & 68.3 & 67.8 \\
SETOKIM \scriptsize{\textcolor{gray}{\textup{[ICLR25]}}} \cite{wu2024towards} & LLaMA2-7B & \ding{51} & - & - & - & 68.0 & 72.4 & 61.2 & 71.3 & 71.3 & - \\
SegLLM \scriptsize{\textcolor{gray}{\textup{[ICLR25]}}} \cite{wang2025segllm} & Vicuna-7B & \ding{51} & 80.2 & 81.5 & 75.4 & 70.3 & 73.0 & 62.5 & 72.6 & 73.6 & 73.6 \\
SegAgent (FT) \scriptsize{\textcolor{gray}{\textup{[CVPR25]}}} \cite{zhu2025segagent} & Qwen-7B & \ding{51} & 79.7 & 81.4 & 76.6 & 72.5 & 75.8 & 66.9 & 75.1 & 75.2 & 75.4 \\
POPEN \scriptsize{\textcolor{gray}{\textup{[CVPR25]}}} \cite{zhu2025popen} & Vicuna-7B & \ding{51} & 79.3 & 82.0 & 74.1 & 73.1 & 77.0 & 65.1 & 75.4 & 75.6 & 75.2 \\
VistaLLM \scriptsize{\textcolor{gray}{\textup{[CVPR24]}}} \cite{pramanick2024jack} & Vicuna-7B & \ding{55} & 74.5 & 76.0 & 72.7 & 69.1 & 73.7 & 64.0 & 69.0 & 70.9 & 71.2 \\
\rowcolor{lightgray} \namexs$^\dagger$\,(FT) \cite{lan2024text4seg} & InternLM2-7B & \ding{55} & 74.7 & 77.4 & 71.6 & 68.5 & 73.6 & 62.9 & 70.7 & 71.6 & 71.4 \\
\rowcolor{lightgray} \namexs \ (FT) \cite{lan2024text4seg} & InternLM2-7B & \ding{51} & 79.2 & 81.7 & 75.6 & 72.8 & 77.9 & 66.5 & 74.0 & 75.3 & 75.4\\
\rowcolor{lightgray} Text4Seg++$^\ddagger$ & Qwen2-7B & \ding{55} & \second{81.5} & \second{83.6} & \best{79.6} & \best{76.9} & \second{81.2} & \best{71.4} & \best{79.6} & \best{80.4} & \best{79.3}  \\
\rowcolor{lightgray} Text4Seg++ & Qwen2-7B & \ding{51} & \best{81.6} & \best{84.1} & \second{78.9} & \best{76.9} & \best{81.7} & \second{70.9} & \second{78.2} & \second{78.9} & \second{78.9}  \\
\midrule
\multicolumn{12}{c}{\textit{\textbf{Generalist Segmentation Models ($\geq$13B)}}} \\
LISA (FT) \scriptsize{\textcolor{gray}{\textup{[CVPR24]}}} \cite{lai2024lisa} & Vicuna-13B & \ding{51} & 76.0 & 78.8 & 72.9 & 65.0 & 70.2 & 58.1 & 69.5 & 70.5 & 70.1 \\
GSVA (FT) \scriptsize{\textcolor{gray}{\textup{[CVPR24]}}} \cite{xia2024gsva} & Vicuna-13B & \ding{51} & 78.2 & 80.4 & 74.2 & 67.4 & 71.5 & 60.9 & 74.2 & 75.6 & 72.8 \\
M$^2$SA \scriptsize{\textcolor{gray}{\textup{[ICLR25]}}} \cite{jang2025mmr} & LLaMA2-13B & \ding{51} & 74.6 & 77.6 & 71.0 & 64.0 & 68.1 & 57.6 & 69.0 & 69.3 & 68.9 \\
VistaLLM \scriptsize{\textcolor{gray}{\textup{[CVPR24]}}} \cite{pramanick2024jack} & Vicuna-13B & \ding{55} & 77.2 & 78.7 & 73.9 & 71.8 & 74.4 & 65.6 & 69.8 & 71.9 & 72.3 \\
\rowcolor{lightgray} \namexs\ (FT) \cite{lan2024text4seg} & Vicuna-13B & \ding{51} & 80.2 & 82.7 & 77.3 & 73.7 & 78.6 & 67.6 & 74.0 & 75.1 & 76.2 \\
\rowcolor{lightgray} Text4Seg++$^\ddagger$ & Qwen2.5-14B & \ding{55} & \second{81.6} & \second{83.7} & \second{79.3} & \second{77.3} & \second{81.3} & \second{72.5} & \best{80.7} & \best{80.8} & \second{79.7}  \\
\rowcolor{lightgray} Text4Seg++ & Qwen2.5-14B & \ding{51} & \best{82.5} & \best{84.9} & \best{79.5} & \best{77.9} & \best{82.5} & \best{72.6} & \second{79.4} & \second{79.7} & \best{79.9} \\
\bottomrule
\end{tabular}
\vspace{-7mm}
\end{center}
\end{table*}

\subsection{Datasets}
Following prior work in multimodal image segmentation \cite{lai2024lisa, rasheed2024glamm, zhang2024groundhog}, we train \namepp on a diverse collection of datasets.
Specifically, we construct our training data using the method introduced in \Cref{box-wise-semantic-descriptors}, leveraging the following datasets:
\begin{itemize}[leftmargin=*]
\item \textbf{COCO Panoptic Segmentation} \cite{lin2014microsoft} is a comprehensive segmentation dataset that includes 80 \textit{thing} categories (\eg, dogs, cats) and 91 \textit{stuff} categories (\eg, grass, sky). We use the training split containing approximately 118,000 images.
\item \textbf{refCOCO series} includes several single-object referring expression segmentation datasets: RefCLEF, RefCOCO, RefCOCO+~\cite{kazemzadeh2014referitgame}, and RefCOCOg \cite{mao2016generation}. We use the \texttt{train} splits for all datasets.
\item \textbf{grefCOCO} \cite{liu2023gres} is a generalized referring expression segmentation dataset designed for multi-object and no-target segmentation tasks. It consists of 278k expressions, including 80k multi-target and 32k no-target expressions.
\item \textbf{Pix2Cap} \cite{you2025pix2cap} contains approximately 20,000 images and 167,254 captions. We treat the caption corresponding to each mask as a referring expression. We use the training split, which includes 18,212 images.
\item \textbf{ReasonSeg} \cite{lai2024lisa} is a single-target reasoning segmentation dataset comprising 1,218 image-instruction-mask samples. The dataset is divided into train, val, and test splits containing 239, 200, and 779 samples, respectively.
\item \textbf{MUSE} \cite{ren2024pixellm} is a multi-target reasoning segmentation dataset with 246,000 question-answer pairs, averaging 3.7 targets per answer. It is split into 239k training, 2.8k validation, and 4.3k test samples.
\item \textbf{RRSIS\_D} \cite{liu2024rotated} is a large-scale dataset for Remote Sensing Referring Image Segmentation (RRSIS), containing 17,402 image–mask–expression triplets. All images are at a resolution of $800 \times 800$ pixels.
\item \textbf{Earthreason} \cite{li2025segearth} is a geospatial pixel-level reasoning dataset designed to evaluate complex real-world remote sensing scenarios. It consists of 5,434 manually annotated image-mask pairs and over 30,000 implicit question–answer pairs, covering 28 scene categories.
\item  \textbf{LLaVA-665k} \cite{liu2024improved} is a visual instruction-following dataset that contains 665k multimodal samples designed to enhance vision–language reasoning capabilities.
\end{itemize}

We summarize the statistics of our training data in \Cref{tab:statistics}.
To advance toward a general-purpose image segmentation framework, we train \namepp on the unified benchmark for 50k steps and evaluate its performance across diverse downstream tasks, \textbf{without any task-specific fine-tuning}.

\begin{table*}[t]
\caption{\footnotesize\\ \textbf{Generalized Referring Expression Segmentation} results on the \textbf{grefCOCO} \cite{liu2023gres} benchmark. 
$^\dagger$ Model based on the 32$\times$ 32 I-SD without the mask refiner.
$^\ddagger$ Model based on the 64$\times$ 64 B-SD without the mask refiner.} 
\label{tab:gres}
\begin{center}
\tabcolsep11pt
\begin{tabular}{llcccccccc}
\toprule
\multirow{2}*{\textbf{Method}} & \multirow{2}*{\textbf{LLM}} & \multirowcell{2}{\textbf{Mask} \\ \textbf{Dec.}} & \multicolumn{2}{c}{\textbf{Validation Set}} & \multicolumn{2}{c}{\textbf{Test Set A}} & \multicolumn{2}{c}{\textbf{Test Set B}} & \multirow{2}*{\textbf{Avg.}}\\
& & & \textbf{gIoU} & \textbf{cIoU} & \textbf{gIoU} & \textbf{cIoU} & \textbf{gIoU} & \textbf{cIoU}  \\
\midrule
\multicolumn{10}{c}{\textit{\textbf{Specialised Segmentation Models}}} \\
ReLA \scriptsize{\textcolor{gray}{\textup{[CVPR23]}}} \cite{liu2023gres} & BERT & \ding{51} &  63.6 & 62.4 & 70.0 & 69.3 & 61.0 & 59.9 & 64.4 \\
LAVT \scriptsize{\textcolor{gray}{\textup{[TPAMI24]}}} \cite{yang2024language} & BERT & \ding{51}  & 58.4 & 57.6 & 65.9 & 65.3 & 55.8 & 55.0 & 59.7 \\
\midrule
\multicolumn{10}{c}{\textit{\textbf{Generalist Segmentation Models (7B)}}} \\
LISA (FT) \scriptsize{\textcolor{gray}{\textup{[CVPR24]}}} \cite{lai2024lisa} & Vicuna-7B & \ding{51}& 61.6 & 61.8 & 66.3 & 68.5 & 58.8 & 60.6 & 62.9 \\
GSVA (FT) \scriptsize{\textcolor{gray}{\textup{[CVPR24]}}} \cite{xia2024gsva} & Vicuna-7B & \ding{51} & 66.5 & 63.3 & 71.1 & 69.9 & 62.2 & 60.5 & 65.6 \\
SAM4MLLM \scriptsize{\textcolor{gray}{\textup{[ECCV24]}}} \cite{chen2024sam4mllm} & Vicuna-7B & \ding{51} & 71.9 & 67.8 & \second{74.2} & \second{72.2} & 65.3 & 63.2 & 69.1 \\
\rowcolor{lightgray} \namexs$^\dagger$ (FT) \cite{lan2024text4seg} & InternLM2-7B & \ding{55} & 71.8 & 65.6 & 71.2 & 70.0 & 64.2 & 62.5 & 67.6 \\
\rowcolor{lightgray} \namexs \ (FT)\ \cite{lan2024text4seg} & InternLM2-7B & \ding{51} & \best{74.4} & 69.1 & \best{75.1} & \best{73.8} & \best{67.3} & \best{66.6} & \best{71.1} \\
\rowcolor{lightgray} Text4Seg++$^\ddagger$ & Qwen2-7B & \ding{55} & 73.5 & \second{69.3} & 72.7 & 71.9 & \second{65.8} & \second{65.6} & 69.8 \\
\rowcolor{lightgray} \namepp & Qwen2-7B & \ding{51} & \second{73.9} & \best{69.4} & 73.5 & \second{72.2} & 65.7 & 65.3 & \second{70.0} \\
\midrule
\multicolumn{10}{c}{\textit{\textbf{Generalist Segmentation Models ($\geq$13B)}}} \\
LISA (FT) \scriptsize{\textcolor{gray}{\textup{[CVPR24]}}} \cite{lai2024lisa} & Vicuna-13B & \ding{51} & 63.5 & 63.0 & 68.2 & 69.7 & 61.8 & 62.2 & 64.7 \\
GSVA (FT) \scriptsize{\textcolor{gray}{\textup{[CVPR24]}}} \cite{xia2024gsva} & Vicuna-13B & \ding{51} & 68.0 & 64.1 & 71.8 & 70.5 & 63.8 & 61.3 & 66.6 \\
\rowcolor{lightgray} \namexs $^\dagger$ (FT)\ \cite{lan2024text4seg} & Vicuna-13B & \ding{55} & 70.3 & 66.9 & 69.8 & 71.4 & 63.8 & 64.4 & 67.8 \\
\rowcolor{lightgray} \namexs \ (FT)\ \cite{lan2024text4seg} & Vicuna-13B & \ding{51} & \best{74.8} & \best{69.8} & \best{75.1} & \best{74.3} & \best{68.0} & \best{67.1} & \best{71.5}  \\
\rowcolor{lightgray} Text4Seg++$^\ddagger$ & Qwen2.5-14B & \ding{55} & 73.4 & \second{69.1} & 72.2 & 71.1 & 65.7 & 65.2 & 69.5  \\
\rowcolor{lightgray} \namepp & Qwen2-7B & \ding{51} & \second{74.1} & \best{69.8} & \second{73.5} & \second{72.1} & \second{65.9} & \second{65.5} & \second{70.2} \\
\bottomrule
\end{tabular}
\end{center}
\end{table*}

\begin{table}[t]
\caption{\footnotesize\\ \textbf{Reasoning Segmentation} results on the \textbf{ReasonSeg} \cite{lai2024lisa} benchmark. 
}
\label{tab:reasoning_segmentation}
\begin{center}
\tabcolsep3pt
\begin{tabular}{llccccc}
\toprule
\multirow{2}*{\textbf{Method}} & \multirow{2}*{\textbf{LLM}} & \multicolumn{2}{c}{\textbf{Val}} & \multicolumn{2}{c}{\textbf{Test}} & \multirow{2}*{\textbf{Avg.}} \\
& & \textbf{gIoU} & \textbf{cIoU} & \textbf{gIoU} & \textbf{cIoU} & \\
\midrule
LISA \scriptsize{\textcolor{gray}{\textup{[CVPR24]}}} \cite{lai2024lisa} & Vicuna-7B & 53.6 & 52.3 & 48.7 & 48.8 & 50.9 \\
SegLLM \scriptsize{\textcolor{gray}{\textup{[ICLR25]}}} \cite{wang2025segllm} & Vicuna-7B & 57.2 & \best{54.3} & 52.4 & 48.4 & 53.1 \\
Seg-Zero \scriptsize{\textcolor{gray}{\textup{[Arxiv25]}}} \cite{liu2025seg} & Qwen2.5-7B & \best{61.6} & \second{52.6} & \best{58.2} & \best{52.4} & \best{56.2} \\
\rowcolor{lightgray} \namepp & Qwen2-7B & \second{59.1} & 49.5 & \second{57.1} & \second{52.1} & \second{54.5} \\
\bottomrule
\end{tabular}
\end{center}
\end{table}

\begin{table}[t]
\caption{\footnotesize\\ \textbf{Multi-target Reasoning Segmentation} results on the \textbf{MUSE}~\cite{ren2024pixellm} benchmark.}
\label{tab:MUSE}
\begin{center}
\tabcolsep3pt
\begin{tabular}{llccccc}
\toprule
\multirow{2}*{\textbf{Method}} & \multirow{2}*{\textbf{LLM}} & \multicolumn{2}{c}{\textbf{Val}} & \multicolumn{2}{c}{\textbf{Test}} & \multirow{2}*{\textbf{Avg.}} \\
& & \textbf{gIoU} & \textbf{cIoU} & \textbf{gIoU} & \textbf{cIoU} & \\
\midrule
LISA \scriptsize{\textcolor{gray}{\textup{[CVPR24]}}} \cite{lai2024lisa} & Vicuna-7B & 17.2 & 28.8 & 24.4 & 36.5 & 26.7 \\
GSVA \scriptsize{\textcolor{gray}{\textup{[CVPR24]}}} \cite{xia2024gsva} & Vicuna-7B & 38.9 & 40.9 & 44.3 & 54.1 & 44.6\\
PixelLM \scriptsize{\textcolor{gray}{\textup{[CVPR24]}}} \cite{ren2024pixellm} & Vicuna-7B & 41.9 & 48.9 & 44.0 & 57.8 & 48.2 \\
POPEN \scriptsize{\textcolor{gray}{\textup{[CVPR25]}}} \cite{zhu2025popen} & Vicuna-7B & \second{45.4} & \second{55.2} & \second{46.4} & \second{62.9} & \second{52.5} \\
\rowcolor{lightgray} \namepp & Qwen2-7B & \best{70.4} & \best{57.7} & \best{63.2} & \best{63.8} & \best{63.8} \\
\bottomrule
\end{tabular}
\end{center}
\end{table}

\subsection{Main Results}

\subsubsection{Referring Expression Segmentation}
We conduct a comprehensive evaluation of our methods on the RefCOCO family of benchmarks and present comparative results in \Cref{tab:res}.
Among 7B-scale discriminative models equipped with mask decoders, GLaMM achieves an average performance of 75.6 cIoU across eight evaluation splits, closely followed by POPEN, which utilizes preference-based optimization and obtains 75.2 cIoU.
SAM4MLLM also demonstrates strong results, achieving the second-best performance on 5 out of 7 individual splits, and ranks second overall in average performance.
For 7B-scale generative models that operate without mask decoders, our \name achieves an average of 71.4 cIoU, marginally outperforming VistaLLM (71.2 cIoU), which generates polygon coordinates to represent segmentation masks.
Remarkably, our proposed \namepp, even without any mask refinement, achieves a significantly higher average of 79.3 cIoU, outperforming the second-best model (76.4 cIoU) by nearly 3 points.
\namepp delivers the best performance on all eight evaluation splits, establishing a new state-of-the-art among both discriminative and generative models at this scale.
When scaled to 13B MLLMs, both \namepp and \name continue to demonstrate strong generalization.
\namepp achieves an average of 79.7 cIoU, while \name follows closely with 76.2 cIoU, both significantly surpassing other competing models, including the generative VistaLLM (72.3 cIoU).
These results highlight the effectiveness and scalability of our text-as-mask framework and demonstrate the superiority of \namepp in referring expression segmentation.
By leveraging the box-wise \textmask and next-brick prediction, \namepp delivers a compact yet expressive textual representation that supports scaling up both mask resolution and training data.

\begin{table}[t]
\caption{\footnotesize\\ \textbf{Open Vocabulary Segmentation} results (\textbf{mIoU}) on various image segmentation benchmarks.}
\label{tab:semantic_segmentation}
\begin{center}
\tabcolsep10pt
\begin{tabular}{lccc}
\toprule
\multirow{2}*{\textbf{Method}} & \multirowcell{2}{\textbf{ADE-150} \\ \textbf{mIoU}} & \multirowcell{2}{\textbf{PC-59} \\ \textbf{mIoU}} & \multirowcell{2}{\textbf{PAS-20} \\ \textbf{mIoU}} \\
\\
\midrule
\multicolumn{4}{c}{\textit{\textbf{Specialised Segmentation Models}}} \\
ClearCLIP \scriptsize{\textcolor{gray}{\textup{[ECCV24]}}} \cite{lan2024clearclip} & 16.7 & 35.9 & 80.9  \\
ProxyCLIP \scriptsize{\textcolor{gray}{\textup{[ECCV24]}}} \cite{lan2024proxyclip} & 24.2 & 39.6 & 83.3 \\
MaskCLIP \scriptsize{\textcolor{gray}{\textup{[ICML23]}}} \cite{ding2022open} & 23.7 & 45.9 & - \\
GroupViT \scriptsize{\textcolor{gray}{\textup{[CVPR22]}}} \cite{xu2022groupvit} & 9.2 & 23.4 & 79.7  \\
OVSeg \scriptsize{\textcolor{gray}{\textup{[CVPR23]}}} \cite{liang2023open} & 24.8 & 53.3 & 92.6 \\
SAN \scriptsize{\textcolor{gray}{\textup{[TPAMI23]}}} \cite{xu2023san} & 27.5 & 53.8 & 94.0 \\
\midrule
\multicolumn{4}{c}{\textit{\textbf{Generalist Segmentation Models (7B)}}} \\
LaSagnA \scriptsize{\textcolor{gray}{\textup{[Arxiv24]}}} \cite{wei2024lasagna} & 14.3 & 46.1 & 69.8  \\
\rowcolor{lightgray} \namexs \ \scriptsize{\textcolor{gray}{\textup{[ICLR25]}}} \cite{lan2024text4seg} & \best{16.5} & \best{52.5} & \best{76.5} \\
\bottomrule
\end{tabular}
\end{center}
\end{table}

\begin{table*}[t]
\caption{\footnotesize\\ \textbf{Referring Expression Segmentation} results on the \textbf{RRSIS-D} \cite{liu2024rotated} benchmark.}
\label{tab:RRSIS_D}
\begin{center}
\tabcolsep10pt
\begin{tabular}{llccccccc}
\toprule
\multirow{2}*{\textbf{Method}} & \multirow{2}*{\textbf{LLM}}  & \multicolumn{3}{c}{\textbf{Validation Set}} & \multicolumn{3}{c}{\textbf{Test Set}} & \multirow{2}*{\textbf{Avg.}}\\
& & \textbf{Acc@0.5} & \textbf{gIoU} & \textbf{cIoU} & \textbf{Acc@0.5} & \textbf{gIoU} & \textbf{cIoU}  \\
\midrule
\multicolumn{9}{c}{\textit{\textbf{Specialised Models}}} \\

RMSIN \scriptsize{\textcolor{gray}{\textup{[CVPR24]}}} \cite{liu2024rotated} & BERT & 74.7 & 65.1 & 78.3 & 74.3 & 64.2 & 77.8 & 72.4 \\
LAVT \scriptsize{\textcolor{gray}{\textup{[TPAMI24]}}} \cite{yang2024language} & BERT & 69.5 & 61.5 & 77.6 & 69.5 & 61.0 & 77.2 & 69.4 \\
\midrule
\multicolumn{9}{c}{\textit{\textbf{Generalist Models}}} \\
LISA \scriptsize{\textcolor{gray}{\textup{[CVPR24]}}} \cite{lai2024lisa} & Vicuna-7B & 27.1 & 27.8 & - & 24.5 & 26.8 & - & - \\
PixelLM \scriptsize{\textcolor{gray}{\textup{[CVPR24]}}} \cite{ren2024pixellm} & Vicuna-7B & 33.5 & 33.7 & - & 28.8 & 31.7 & - & - \\
NEXT-Chat \scriptsize{\textcolor{gray}{\textup{[ICML24]}}} \cite{zhang2023next} & Vicuna-7B & 29.0 & 27.0 & - & 26.4 &  25.0 & - & - \\
GeoGround \scriptsize{\textcolor{gray}{\textup{[Arxiv24]}}}  \cite{zhou2024geoground} & Vicuna-7B & 68.7 & 61.1 & - & 67.5 & 60.5 & - & - \\
SegEarth-R1 (FT) \scriptsize{\textcolor{gray}{\textup{[Arxiv25]}}} \cite{li2025segearth} & Phi-1.5-1.3B & \best{78.6} & \best{67.6} & \best{78.9} & \best{77.0} & \best{66.4} & \best{78.0} & \best{74.4} \\
\rowcolor{lightgray} \namepp & Qwen2-7B & \second{74.8} & \second{64.1} & \second{75.8} & \second{73.2} & \second{62.8} & \second{74.2} & \second{70.8} \\
\bottomrule
\end{tabular}
\vspace{-5mm}
\end{center}
\end{table*}

\begin{table}[t]
\caption{\footnotesize\\ \textbf{Geospatial pixel reasoning} results on the \textbf{EarthReason}~\cite{li2025segearth} benchmark.}
\label{tab:geospatial_reasoning_segmentation}
\begin{center}
\tabcolsep3pt
\begin{tabular}{llccccc}
\toprule
\multirow{2}*{\textbf{Method}} & \multirow{2}*{\textbf{LLM}} & \multicolumn{2}{c}{\textbf{Val}} & \multicolumn{2}{c}{\textbf{Test}} & \multirow{2}*{\textbf{Avg.}} \\
& & \textbf{gIoU} & \textbf{cIoU} & \textbf{gIoU} & \textbf{cIoU} & \\
\midrule
\multicolumn{7}{c}{\textit{\textbf{Generalist Models}}} \\
LISA (FT) \cite{lai2024lisa} & Vicuna-7B & 61.0 & 57.4 & 60.9 & 59.1 & 59.6 \\
PixelLM (FT) \cite{ren2024pixellm} & Vicuna-7B & 57.9 & 57.8 & 60.0 & 59.2 & 58.7 \\
SegEarth-R1 (FT) \cite{li2025segearth} & Phi-1.5-1.3B & \second{68.6} & \second{64.1} & \second{70.8} & \best{68.3} & \second{68.0} \\
\rowcolor{lightgray} \namepp & Qwen2-7B & \best{71.9} & \best{69.8} & \best{73.0} & \second{65.6} & \best{70.1} \\
\bottomrule
\end{tabular}
\end{center}
\end{table}

\subsubsection{Generalized Referring Expression Segmentation}
We further evaluate our methods on generalized referring expression segmentation benchmark, which includes both multi-object and no-object cases, as shown in \Cref{tab:gres}.
Without any task-specific design, both \name and \namepp maintain strong performance in this more challenging setting.
At the 7B scale, \namepp without the mask refiner achieves an average score of 69.8, outperforming GSVA (65.6) by over 4 points and exceeding \name without mask refinement (67.6).
However, when equipped with a mask refiner and fine-tuned exclusively on this benchmark, \name achieves a higher average of 71.1, surpassing \namepp.
At the 13B scale, \name further extends its lead, achieving an average of 71.5, outperforming GSVA by 4.9 points, while \namepp records a strong 69.5.
These results underscore the robustness and versatility of our \name and \namepp in handling more complex segmentation scenarios involving multiple or absent referents.

\subsubsection{Reasoning Segmentation}
We further assess the conversational image segmentation capabilities of \namepp on two challenging benchmarks specifically designed to assess reasoning segmentation in vision-language models.
For a fair comparison with all baselines that incorporate mask decoders, we equip \namepp with SAMRefiner as an off-the-shelf post-processing mask refiner.
The first benchmark, ReasonSeg~\cite{lai2024lisa}, focuses on complex reasoning grounded in world knowledge and requires models to handle implicit, compositional, or abstract query texts.
As shown in \Cref{tab:reasoning_segmentation}, \namepp achieves an average score of 54.5, significantly outperforming LISA (50.9), and performing comparably to SegLLM (53.1).
However, it slightly underperforms Seg-Zero (56.2), which benefits from a dedicated reasoning-chain-guided segmentation mechanism tailored for this task.

The second benchmark, MUSE~\cite{ren2024pixellm}, presents a more complex multi-object reasoning segmentation challenge.
As reported in \Cref{tab:MUSE}, \namepp significantly outperforms all existing baselines by a large margin, achieving an average score of 63.8.
This is 11.3 points higher than the best-performing baseline, POPEN (52.5), and dramatically outperforms early models like LISA (26.7), GSVA (44.6), and PixelLLM (48.2).
Built purely on generative language modeling, \namepp achieves superior reasoning ability, segmentation precision, and multi-object understanding without requiring architectural customization.

\subsubsection{Open Vocabulary Segmentation}
We follow LaSagnA \cite{wei2024lasagna} to evaluate the performance of \name on open-vocabulary segmentation tasks.
We evaluate the model's performance on ADE20K (A-150) \cite{zhou2019semantic}, PASCAL Context 59 (PC-59) \cite{mottaghi2014role}, and PASCAL VOC 20 (PAS-20) \cite{everingham2009pascal} datasets, using mIoU as the evaluation metric.

As demonstrated in \Cref{tab:semantic_segmentation},
it is expected that \name falls behind specialized segmentation models (\eg, ProxyCLIP \cite{lan2024proxyclip}, OVSeg \cite{liang2023open}, and SAN \cite{xu2023side}), because LLMs typically require quite large datasets to be sufficiently trained. 
However, \name still demonstrates competitive performance on the PC-59 benchmark, underscoring its efficiency. More importantly, it significantly outperforms the MLLM-based LaSagnA, which uses an additional decoder, showcasing its strong potential for open-vocabulary segmentation.

\subsubsection{Extend to Remote Sensing Image Segmentation}
To further evaluate the generalization ability of \namepp beyond natural images, we conduct experiments on remote sensing image segmentation tasks.
We consider two challenging benchmarks.
On RRSIS-D~\cite{liu2024rotated} shown in \Cref{tab:RRSIS_D}, which focuses on referring expression segmentation in remote sensing scenes, \namepp achieves an average score of 70.8, closely approaching the best-performing specialized models, including RMSIN (72.4).
It underperforms the best-performing specialized model, SegEarth-R1 (74.4), despite being trained with a unified formulation and no domain-specific finetuning.
Compared to previous vision-language models such as GeoGround, \namepp demonstrates significantly better localization and semantic understanding in high-resolution aerial imagery.

EarthReason~\cite{li2025segearth} benchmark is a recently introduced task designed to assess precise geospatial pixel reasoning ability.
As shown in \Cref{tab:geospatial_reasoning_segmentation}, \namepp achieves a new state-of-the-art performance with an average score of 70.1.
This surpasses the strong baseline SegEarth-R1 (68.0) and significantly outperforms PixelLM (58.7) and LISA (59.6).
These results highlight the scalability, domain transferability, and robust reasoning capabilities of our text-as-mask framework in diverse settings, including complex geospatial environments.
Notably, \namepp achieves this performance without any architectural modification or remote-sensing-specific module, reinforcing its potential as a unified segmentation solution across domains.

\begin{table*}[t]
\caption{\footnotesize\\ \textbf{Referring Expression Comprehension} results (\textbf{Acc@0.5}) on \textbf{RefCOCO (+/g)} \cite{kazemzadeh2014referitgame,mao2016generation} benchmarks. 
}
\label{tab:rec}
\begin{center}
\tabcolsep9pt
\begin{tabular}{llccccccccc}
\toprule
\multirow{2}*{\textbf{Method}} & \multirow{2}*{\textbf{LLM}} & \multicolumn{3}{c}{\textbf{refCOCO}} & \multicolumn{3}{c}{\textbf{refCOCO+}} & \multicolumn{2}{c}{\textbf{refCOCOg}} & \multirow{2}*{\textbf{Avg.}}\\
& & \textbf{val} & \textbf{testA} & \textbf{testB} & \textbf{val} & \textbf{testA} & \textbf{testB} & \textbf{val} & \textbf{test} & \\
\midrule
\multicolumn{11}{c}{\textit{\textbf{Specialised Models}}} \\
PolyFormer-L \scriptsize{\textcolor{gray}{\textup{[CVPR23]}}} \cite{liu2023polyformer} & BERT & 90.4 & 92.9 & 87.2 & 85.0 & 89.8 & 78.0 & 85.8 & 85.9 & 86.9 \\
UNINEXT-L \scriptsize{\textcolor{gray}{\textup{[CVPR24]}}}
\cite{yan2023universal} & BERT & 91.4 & 93.7 & 88.9 & 83.1 & 87.9 & 76.2 & 86.9 & 87.5 & 87.0 \\
G-DINO \scriptsize{\textcolor{gray}{\textup{[ECCV24]}}} \cite{liu2024grounding} & BERT & 90.6 & 93.2 & 88.2 & 82.8 & 89.0 & 75.9 & 86.1 & 87.0 & 86.6 \\
\midrule
\multicolumn{11}{c}{\textit{\textbf{Generalist Models (7B)}}} \\
Shikra \scriptsize{\textcolor{gray}{\textup{[Arxiv23]}}} \cite{chen2023shikra} & Vicuna-7B & 87.0 & 90.6 & 80.2 & 81.6 & 87.4 & 72.1 & 82.3 & 82.2 & 82.9\\
Qwen2-VL \scriptsize{\textcolor{gray}{\textup{[Arxiv24]}}} \cite{wang2024qwen2} & Qwen2-7B & 91.7 & 93.6 & 87.3 & 85.8 & 90.5 & 79.5 & 87.3 & 87.8 & 87.9 \\
Qwen2.5-VL \scriptsize{\textcolor{gray}{\textup{[Arxiv24]}}} \cite{bai2025qwen2} & Qwen2.5-7B & 90.0 & 92.5 & 85.4 & 84.2 & 89.1 & 76.9 & 87.2 & 87.2 & 86.6 \\
InternVL2 \scriptsize{\textcolor{gray}{\textup{[Arxiv24]}}} \cite{chen2024far} & InternLM2-7B & 87.1 & 91.1 & 80.7 & 79.8 & 87.9 & 71.4 & 82.7 & 82.7 & 82.9 \\
InternVL3 \scriptsize{\textcolor{gray}{\textup{[Arxiv25]}}} \cite{zhu2025internvl3} & Qwen2.5-7B & \second{92.5} & \second{94.6} & 88.0 & \second{88.2} & \second{92.5} & \second{81.8} & \second{89.6} & \second{90.0} & \second{89.7} \\
LISA \scriptsize{\textcolor{gray}{\textup{[CVPR24]}}} \cite{lai2024lisa} & Vicuna-7B & 85.4 & 88.8 & 82.6 & 74.2 & 79.5 & 68.4 & 79.3 & 80.4 & 79.8 \\
GSVA \scriptsize{\textcolor{gray}{\textup{[CVPR24]}}} \cite{xia2024gsva} & Vicuna-7B & 86.3 & 89.2 & 83.8 & 72.8 & 78.8 & 68.0 & 81.6 & 81.8 & 80.3 \\
VistaLLM \scriptsize{\textcolor{gray}{\textup{[CVPR24]}}} \cite{pramanick2024jack} & Vicuna-7B & 88.1 & 91.5 & 83.0 & 82.9 & 89.8 & 74.8 & 83.6 & 84.4 & 84.8 \\
Groma \scriptsize{\textcolor{gray}{\textup{[ECCV24]}}} \cite{ma2024groma} & Vicuna-7B & 89.5 & 92.1 & 86.3 & 83.9 & 88.9 & 78.1 & 86.4 & 87.0 & 86.5 \\
SegLLM \scriptsize{\textcolor{gray}{\textup{[ICLR25]}}} \cite{wang2025segllm} & Vicuna-7B & 90.0 & 92.1 & 86.2 & 82.2 & 85.5 & 76.1 & 83.9 & 85.9 & 85.2 \\
\rowcolor{lightgray} \namexs \ \cite{lan2024text4seg} & InternLM2-7B & 90.3 & 93.4 & 87.5 & 85.2 & 89.9 & 79.5 & 85.4 & 85.4 & 87.1\\
\rowcolor{lightgray} \namepp & Qwen2-7B & \best{93.2} & \best{95.3} & \best{90.7} & \best{89.7} & \best{93.2} & \best{84.5} & \best{90.8} & \best{91.2} & \best{91.1} \\
\midrule
\multicolumn{11}{c}{\textit{\textbf{Generalist Models ($\geq$13B)}}} \\
Shikra \scriptsize{\textcolor{gray}{\textup{[Arxiv23]}}} \cite{chen2023shikra} & Vicuna-13B & 87.8 & 91.1 & 81.8 & 82.9 & 87.8 & 74.4 & 82.6 & 83.2 & 84.0\\
LISA \scriptsize{\textcolor{gray}{\textup{[CVPR24]}}} \cite{lai2024lisa} & LLaMA2-13B & 85.9 & 88.8 & 81.7 & 74.5 & 80.6 & 68.3 & 80.1 & 81.3 & 80.2 \\
GSVA \scriptsize{\textcolor{gray}{\textup{[CVPR24]}}} \cite{xia2024gsva} & LLaMA2-13B & 89.2 & 92.1 & 87.2 & 79.7 & 84.5 & 73.4 & 85.5 & 86.2 & 84.7 \\
VistaLLM \scriptsize{\textcolor{gray}{\textup{[CVPR24]}}} \cite{pramanick2024jack} & Vicuna-13B & 89.9 & 92.5 & 85.0 & 84.1 & 90.3 & 75.8 & 86.0 & 86.4 & 86.3 \\
InternVL3 \scriptsize{\textcolor{gray}{\textup{[Arxiv25]}}} \cite{zhu2025internvl3} & Qwen2.5-14B & \second{92.0} & \second{94.4} & 87.8 & \second{87.4} & \second{92.1} & \second{81.5} & \second{88.6} & \second{89.3} & \second{89.1} \\
\rowcolor{lightgray} \namexs \ \cite{lan2024text4seg} & Vicuna-13B & 91.2 & 94.3 & \second{88.0} & 85.7 & 90.8 & 80.1 & 85.6 & 85.5 & 87.7 \\
\rowcolor{lightgray} \namepp & Qwen2.5-14B & \best{94.3} & \best{96.2} & \best{91.4} & \best{90.6} & \best{94.2} & \best{86.1} & \best{91.6} & \best{91.8} & \best{92.0} \\
\bottomrule
\end{tabular}
\end{center}
\end{table*}

\begin{table*}[t]
\caption{\footnotesize\\ \textbf{Visual Question Answering} and \textbf{Referring Expression Segmentation} results on various benchmarks. 
Mix$^\dagger$ is a combination of referring segmentation, semantic segmentation and VQA datasets from LISA.}
\label{tab:vqa}
\begin{center}
\tabcolsep3pt
\begin{tabular}{llc|cccccc|ccc}
\toprule
\multirow{2}*{\textbf{Method}} & \multirow{2}*{\textbf{LLM}} & \multirow{2}*{\textbf{Training Data}} & \multicolumn{6}{c|}{\textbf{VQA}} & \multicolumn{3}{c}{\textbf{RES (val)}} \\
& & & \textbf{VQAv2} & \textbf{GQA} & \textbf{VisWiz} & \textbf{ScienceQA} & \textbf{TextQA} & \textbf{POPE} & \textbf{refCOCO} & \textbf{refCOCO+} & \textbf{refCOCOg} \\
\midrule
LISA \scriptsize{\textcolor{gray}{\textup{[CVPR24]}}} \cite{lai2024lisa} & Vicuna-7B & Mix$^\dagger$ & - & - & - & - & - & - & 74.1 & 62.4 & 66.4 \\
LLaVA-1.5 \scriptsize{\textcolor{gray}{\textup{[CVPR24]}}} \cite{liu2024improved} & Vicuna-7B & 665k & 78.0 & 61.7 & 50.6 & 68.4 & 55.0 & 85.4 & - & - & - \\
\rowcolor{lightgray} \namexs \ \cite{lan2024text4seg} & Vicuna-7B & 665k + refseg & 76.6 & 60.2 & 50.9 & 68.1 & 55.0 & 84.2 & 77.5 & 70.7 & 73.4 \\
\bottomrule
\end{tabular}
\vspace{-3mm}
\end{center}
\end{table*}

\subsubsection{Referring Expression Comprehension}
As \namepp leverages the box-wise \textmask formulation, it naturally unifies visual grounding and segmentation within a single, compact representation.
To assess its capacity for referring expression comprehension, we evaluate the model’s ability to localize referred objects using bounding boxes, measured by the standard Acc@0.5 metric.
Experiments are conducted on the widely used RefCOCO benchmarks.
As shown in \Cref{tab:rec}, \namepp achieves superior performance across all evaluation splits compared to both specialized and generalist baselines.
At the 7B scale, \namepp sets a new state-of-the-art with an average Acc@0.5 of 91.1, outperforming the strongest prior model, InternVL3 (89.7).
\namepp consistently surpasses recent generalist MLLMs such as Qwen2-VL (87.9), VistaLLM (84.8), and SegLLM (85.2)
Specifically, under 7B scale, \namepp obtains the best results on all evaluation splits, with an average of 91.1 that is significantly higher than existing SOTA method, \eg, InternVL3 at 89.7.
At the $\geq$13B scale, \namepp further improves to an average score of 92.0, again outperforming prior best models such as InternVL3-14B (89.1) and GSVA (84.7).
Notably, it achieves the best results on all splits, including 86.1 on RefCOCO+ testB, compared to 81.5 by InternVL3.
These results highlight a key advantage of our unified formulation: by jointly modeling “where to look” and “what to segment”, \namepp benefits from dense mask supervision, which strengthens spatial understanding and significantly enhances grounding precision.

\subsubsection{Visual Understanding}
Our text-as-mask paradigm allows for seamless integration of downstream segmentation task into the pre-training of MLLMs. 
To evaluate its effectiveness, we assess the model's performance on various visual understanding benchmarks, using the LLaVA-1.5-7B model as the baseline. 
Our method, \name, built upon the stage-2 of LLaVA-1.5-7B, is trained on both the LLaVA-v1.5-mix665k dataset and our referring segmentation datasets. 
For a comprehensive comparison, we also report the performance of the LLaVA-1.5-7B model based on our implementation.

\Cref{tab:vqa} presents a comprehensive comparison between LLaVA-1.5 and \name across various VQA and RES benchmarks.
Notably, \name, trained on a mixed dataset, achieves performance on par with LLaVA-1.5 in visual question answering tasks while delivering strong results on RES benchmarks.
These results validate that our text-as-mask based segmentation method acts as a seamless enhancement, offering a streamlined approach for pre-training MLLMs. 
It successfully integrates robust segmentation functionality without compromising the model’s conversational capabilities.

\begin{figure}[t]
    \centering
    \includegraphics[width=1.0\linewidth]{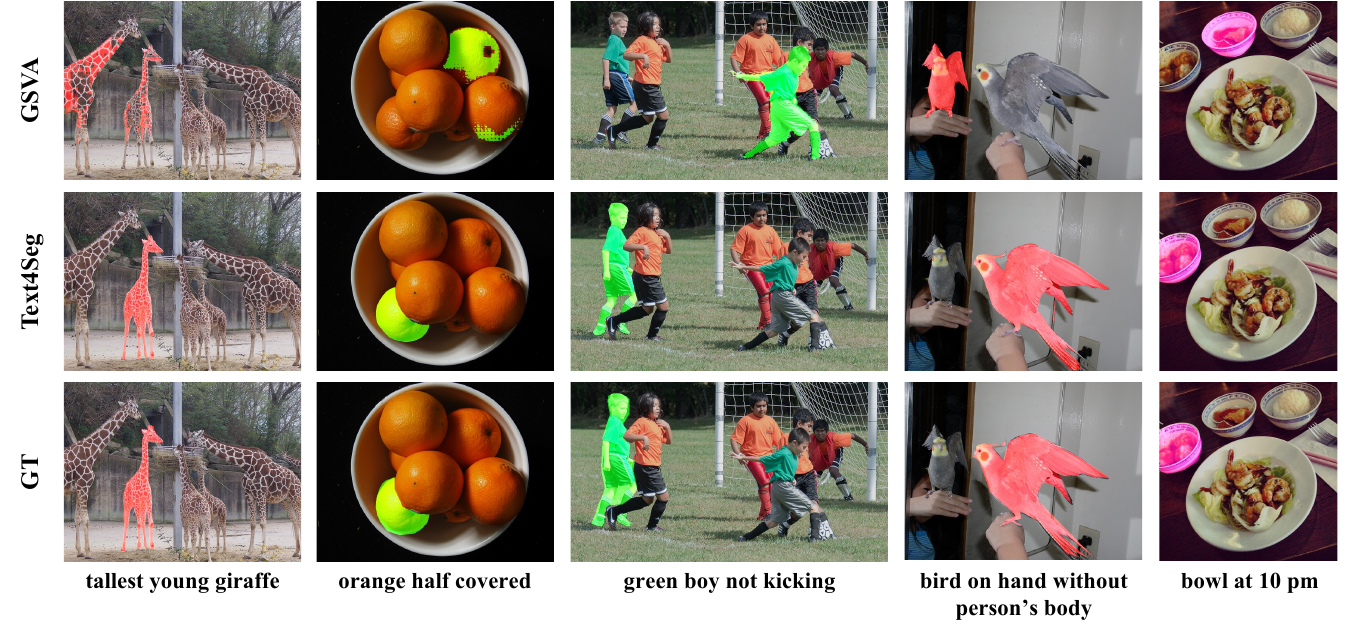}
    \captionsetup{font={footnotesize}}
    \caption{Qualitative results of \name and GSVA \cite{xia2024gsva} on the RES task. The corresponding referring expressions are displayed in the bottom. }
    \label{fig:res_vis}
\end{figure}

\begin{figure}[t]
    \centering
    \includegraphics[width=1.0\linewidth]{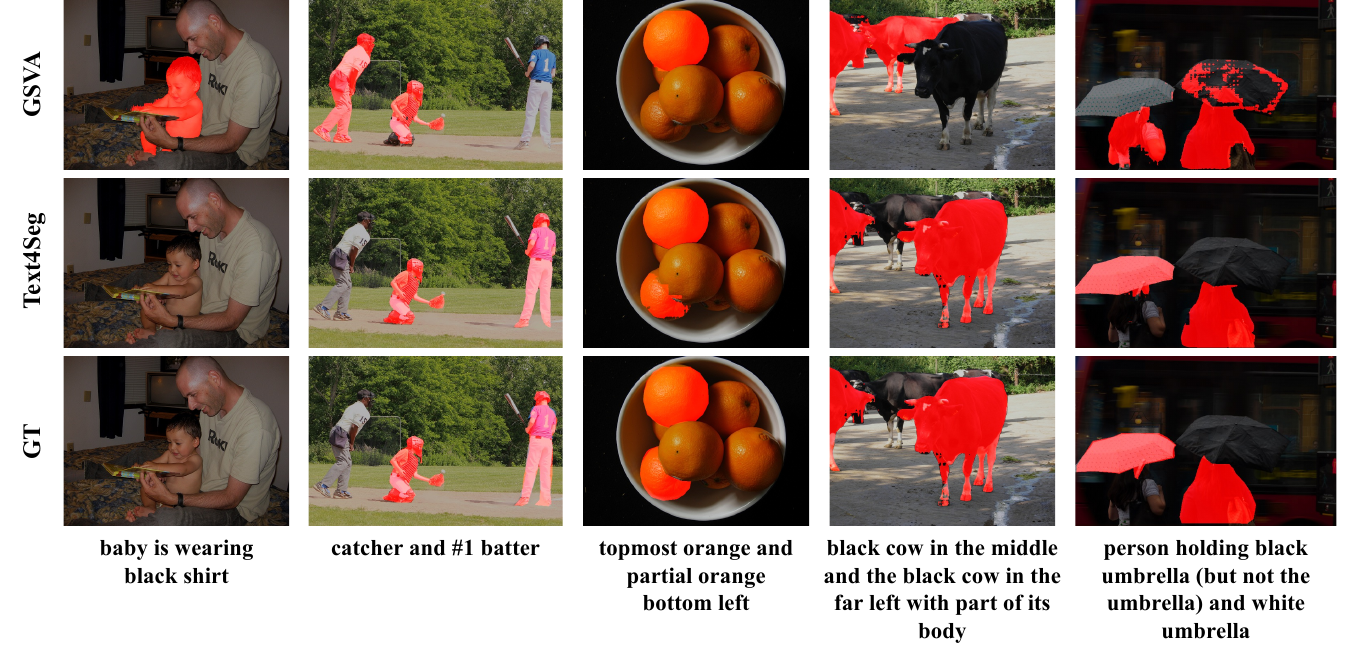}
    \captionsetup{font={footnotesize}}
    \caption{Qualitative results of \name and GSVA \cite{xia2024gsva} on the GRES task.}
    \label{fig:gres_vis}
\end{figure}

\begin{figure}[t]
    \centering
    \includegraphics[width=1.0\linewidth]{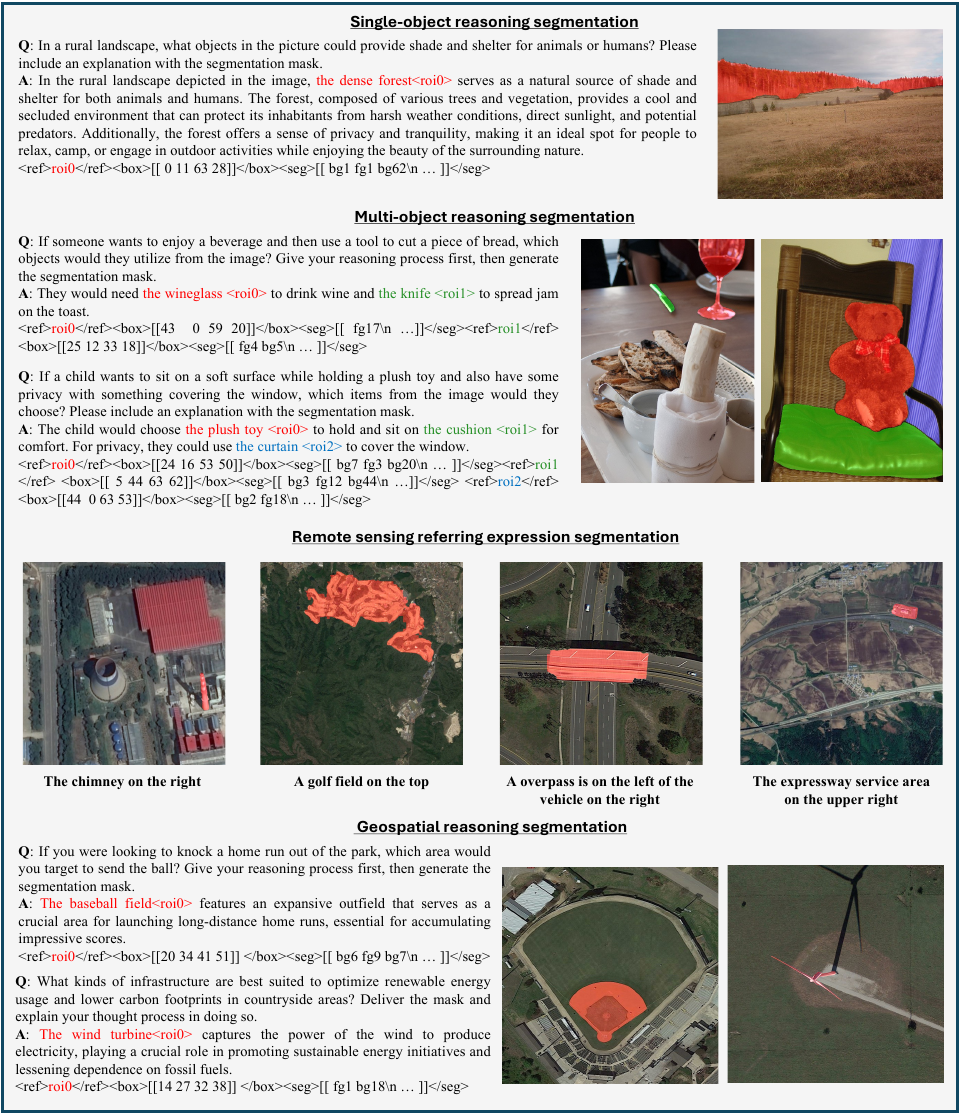}
    \captionsetup{font={footnotesize}}
    \caption{Qualitative results of \namepp across various vision-language tasks and diverse scenarios,
    including challenging tasks on remote sensing datasets.}
    \label{fig:visual_manytasks}
\end{figure}

\begin{figure*}[t]
\centering
\includegraphics[width=0.9\textwidth]{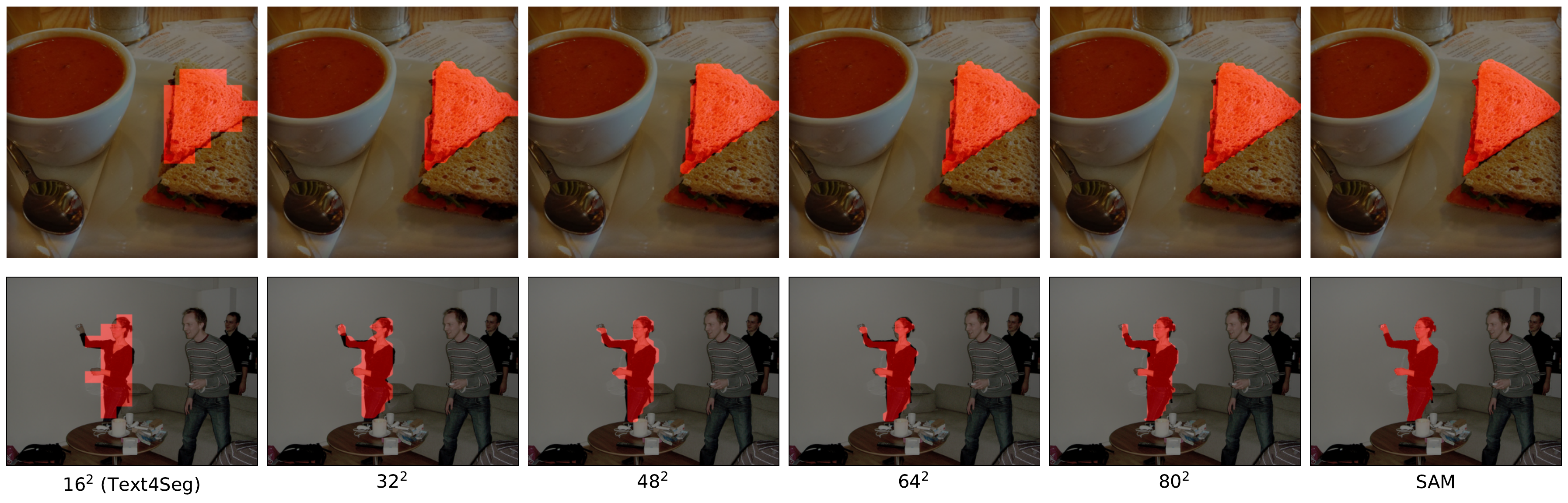}
\captionsetup{font={footnotesize}}
\caption{Visualization of RES results across different resolutions, and with SAM as mask refiner.}
\label{fig:resolution_vis}
\end{figure*}

\begin{figure}[t]
    \centering
    \includegraphics[width=0.95\linewidth]{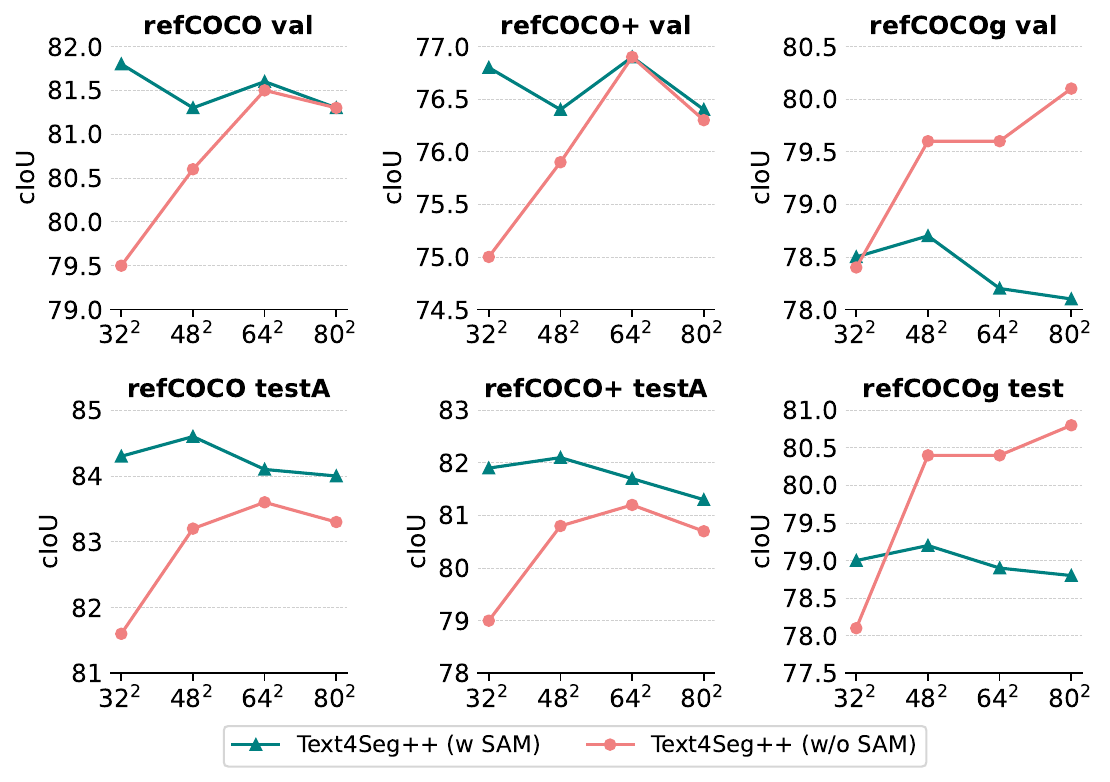}
    \captionsetup{font={footnotesize}}
    \caption{Quantitative comparison of referring expression segmentation results using different resolutions of box-wise \textmask.}
    \label{fig:output_resolutions_vis}
\end{figure}

\begin{figure}[t]
    \centering
    \includegraphics[width=0.65\linewidth]{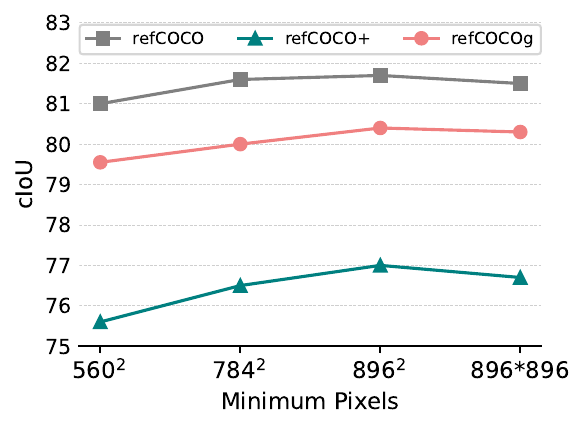}
    \captionsetup{font={footnotesize}}
    \caption{Quantitative comparison of referring expression segmentation results with different image input resolutions. All images are resized to meet a specified minimum pixel count before being fed into the model.}
    \label{fig:input_resolutions_vis}
\end{figure}

\begin{figure}[t]
    \centering
    \includegraphics[width=0.95\linewidth]{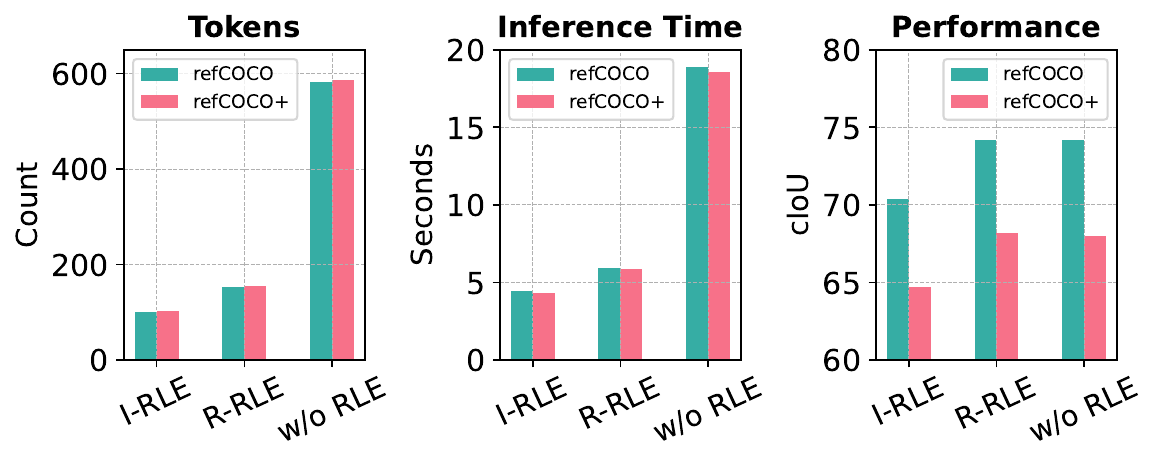}
    \captionsetup{font={footnotesize}}
    \caption{Comparison of different encoding strategies for \textmask at the $16 \times 16$ resolution. Experiments are conducted on a standard NVIDIA V100 GPU. Our proposed Row-wise Run-Length Encoding (\rle) achieves an optimal trade-off between token efficiency, inference speed, and accuracy.}
    \label{fig:rle}
    \vspace{-2mm}
\end{figure}

\subsection{Visualization Analysis}

We present qualitative comparisons between \name and GSVA in \Cref{fig:res_vis,fig:gres_vis} to highlight the effectiveness of our text-as-mask framework across different segmentation scenarios.
In the single-object RES task, \name demonstrates a superior understanding of referring expressions, generating more accurate and precise segmentation maps compared to GSVA.
In the GRES task (\Cref{fig:gres_vis}), GSVA tends to incorrectly segment empty objects despite the inclusion of a \texttt{$<$REJ$>$} token (as seen in the first two columns). 
In contrast, \name consistently avoids such mistakes by labeling them as ``\texttt{others}" without special design.
Furthermore, \name significantly outperforms GSVA in the multiple-object RES task, delivering more precise segmentation results with better grounding performance.
These results fully validate the effectiveness of \name in handling diverse and challenging visual grounding and segmentation tasks.

To further assess generalization and task versatility, we visualize qualitative results from \namepp across a wide range of vision-language segmentation tasks in \Cref{fig:visual_manytasks}.
These examples demonstrate that our enhanced model with box-wise \textmask can effectively handle tasks that go beyond explicit referring expressions. 
For instance, in single-object reasoning segmentation from the ReasonSeg dataset and multi-object reasoning tasks from the MUSE benchmark, \namepp accurately segments semantically relevant regions using abstract region tags like ``\texttt{roi0, roi1, ...}", showcasing its capacity for implicit reasoning and compositional understanding.
Moreover, in remote sensing tasks (last two rows), \namepp exhibits strong domain generalization, producing precise masks for complex aerial imagery.
These results collectively highlight the versatility, robustness, and fine-grained segmentation capabilities of \namepp across both natural and geospatial domains, validating its unified and generative formulation for a broad spectrum of vision-language segmentation tasks.

\subsection{Ablation Study and Analysis}


\subsubsection{Resolution of Semantic Descriptors}
To assess the impact of semantic descriptor granularity on segmentation performance, we construct instruction-tuning datasets using box-wise \textmask at varying spatial resolutions, ranging from $32 \times 32$ to $80 \times 80$. 
For reference, we also include the $16 \times 16$ image-wise \textmask configuration used in \name.
As illustrated in \Cref{fig:resolution_vis}, higher-resolution \textmask representations yield qualitatively finer segmentation outputs. 
Increasing the resolution leads to more precise object boundaries and improved structural detail. 
In contrast, lower-resolution settings, such as $16 \times 16$, exhibit blocky artifacts and under-segmentation due to coarse spatial encoding. 
Notably, the 64$^2$ or 80$^2$ configuration achieves segmentation quality that closely resembles results obtained with an external mask refiner like SAM.

These observations are quantitatively validated in \Cref{fig:output_resolutions_vis}, where we report average cIoU scores across all splits of the RefCOCO, RefCOCO+, and RefCOCOg benchmarks. 
The results show a consistent improvement in performance as the resolution of the box-wise \textmask increases. 
Importantly, the $64 \times 64$ resolution already achieves parity, or even slightly outperforms, the SAM-refined variant of our model, indicating that dense and compact semantic tokens can drive high-quality segmentation in a fully generative manner.
Given its strong balance between performance and sequence length, we adopt the $64 \times 64$ resolution as the default setting in our experiments.

\subsubsection{Resolution of Input Image}
Unlike most methods that leverage SAM's image encoder, typically operating at high resolutions such as 1024$^2$, paired with a mask decoder,  \namepp performs segmentation solely through the vision-language modeling capabilities of MLLMs. 
As a result, its ability to perceive fine-grained visual details is directly influenced by the resolution of the input image, making it an important factor to investigate.
To assess this impact, we evaluate \namepp across four different input resolutions.
As shown in \Cref{fig:input_resolutions_vis}, we report the average cIoU scores over the refCOCO, refCOCO+, and refCOCOg benchmarks.
The results exhibit a clear trend: segmentation performance improves consistently as the input resolution increases, highlighting the critical role of high-resolution visual input in enhancing spatial understanding and mask quality.
In particular, increasing resolution from 560$^2$ to 896$^2$ (while preserving the original aspect ratio) yields a notable accuracy boost.
However, forcing the image into a square shape (896 * 896) introduces slight distortions and results in a marginal performance drop.
Based on this trade-off between spatial fidelity and computational efficiency, we adopt 784$^2$ (minimum pixels with the native aspect ratio) as the default input resolution in all experiments.


\begin{table}[t]
\caption{\footnotesize \\ Ablation results of \namepp with and without semantic bricks (SB). Results are reported as the average mIoU across all splits of each dataset.}
\label{tab:semantic_bricks}
\begin{center}
\tabcolsep6pt
\begin{tabular}{lcccc}
\toprule
\textbf{Method} & SB & \textbf{refCOCO} & \textbf{refCOCO+} & \textbf{refCOCOg} \\
\midrule
\namepp & \ding{55} & 79.7 & 74.9 & 78.1 \\
\rowcolor{lightgray} \namepp & \ding{51} & 79.7 & 75.1 & 78.3 \\
\bottomrule
\end{tabular}
\end{center}
\end{table}

\subsubsection{I-RLE v.s. \rle}

We investigate the impact of different encoding methods for \textmask at a $16\times16$ resolution using the \texttt{train/val} splits of the refCOCO and refCOCO+ datasets. 
As illustrated in \Cref{fig:rle}, while full-length \textmask achieve high performance, they suffer from significantly longer inference times ($\sim$19 seconds) due to longer output tokens ($\sim$590) on both datasets. 
Although the I-RLE method reduces both the number of tokens and inference time, it results in a notable performance drop, from 74.2 to 70.4 cIoU on refCOCO and 68.0 to 64.7 cIoU on refCOCO+. 
Our proposed \rle method strikes a better balance, reducing the length of \textmask by 74\% and improving inference speed by an average of $3\times$, while still maintaining nearly the same performance.
These results highlight the effectiveness of \rle as an efficient yet lossless encoding mechanism for integrating dense segmentation into the generative pipeline of MLLMs.

\subsubsection{Ablation study about Semantic Bricks}

We assess the impact of Semantic Bricks (SB) by comparing \namepp\ with and without this component using the 32$\times$32 box-wise semantic descriptors.
As shown in Table~\ref{tab:semantic_bricks}, removing SB yields minimal changes in performance, but slightly degrades results on refCOCO+ and refCOCOg. 
This suggests that our box-wise semantic descriptor with semantic bricks provide minor yet consistent benefits.
More importantly, as illustrated in \Cref{fig:tokens_resolution}, Semantic Bricks significantly reduce the sequence length, enabling the use of higher-resolution B-SD (64$\times$64) for finer-grained segmentation while also accelerating both training and inference.

\section{Discussion}\label{sec:conclusion}
\subsection{Conclusion}
In this work, we presented text-as-mask, a novel paradigm that fundamentally recasts image segmentation as a text generation problem within Multimodal Large Language Models (MLLMs). This approach eliminates the need for additional decoders, simplifying the integration of dense prediction tasks. 
Our initial image-wise \textmask provided a patch-aligned textual representation, enhanced by Row-wise Run-Length Encoding (RLE) for significant length reduction and inference speedup, forming the foundation of our \name framework.
Building upon this, we further refined our approach with box-wise \textmask.
This more compact, region-level representation leverages bounding boxes and semantic bricks, leading to our next-brick prediction framework: \namepp.
\namepp not only achieves significantly finer-grained segmentation compared to its predecessor but also maintains a reduced sequence length, showcasing a powerful combination of precision, scalability, and generative efficiency.
Extensive experiments across a wide range of benchmarks, including referring expression, reasoning, and challenging tasks like remote sensing segmentation, consistently demonstrated that \namepp surpasses existing state-of-the-art methods. 
Importantly, it achieves this superior performance without any task-specific fine-tuning or architectural modifications, showcasing its exceptional versatility and robustness across diverse domains and tasks.

\subsection{Future works and broader impact}
Our work underscores the potential of treating dense prediction as a generative language modeling task. 
In the future, we believe that the \textit{text-as-mask} paradigm opens promising new research directions for integrating fine-grained visual understanding into large-scale vision-language models in a principled, efficient, and generalizable manner.



\ifCLASSOPTIONcaptionsoff
  \newpage
\fi



\bibliographystyle{IEEEtran}
\bibliography{IEEEexample}

\begin{thebibliography}{10}
\providecommand{\url}[1]{#1}
\csname url@samestyle\endcsname
\providecommand{\newblock}{\relax}
\providecommand{\bibinfo}[2]{#2}
\providecommand{\BIBentrySTDinterwordspacing}{\spaceskip=0pt\relax}
\providecommand{\BIBentryALTinterwordstretchfactor}{4}
\providecommand{\BIBentryALTinterwordspacing}{\spaceskip=\fontdimen2\font plus
\BIBentryALTinterwordstretchfactor\fontdimen3\font minus \fontdimen4\font\relax}
\providecommand{\BIBforeignlanguage}[2]{{%
\expandafter\ifx\csname l@#1\endcsname\relax
\typeout{** WARNING: IEEEtran.bst: No hyphenation pattern has been}%
\typeout{** loaded for the language `#1'. Using the pattern for}%
\typeout{** the default language instead.}%
\else
\language=\csname l@#1\endcsname
\fi
#2}}
\providecommand{\BIBdecl}{\relax}
\BIBdecl

\bibitem{yin2024survey}
S.~Yin, C.~Fu, S.~Zhao, K.~Li, X.~Sun, T.~Xu, and E.~Chen, ``A survey on multimodal large language models,'' \emph{National Science Review}, vol.~11, no.~12, 2024.

\bibitem{liu2024visual}
H.~Liu, C.~Li, Q.~Wu, and Y.~J. Lee, ``Visual instruction tuning,'' \emph{Advances in neural information processing systems}, vol.~36, 2024.

\bibitem{lu2024deepseek}
H.~Lu, W.~Liu, B.~Zhang, B.~Wang, K.~Dong, B.~Liu, J.~Sun, T.~Ren, Z.~Li, H.~Yang, Y.~Sun, C.~Deng, H.~Xu, Z.~Xie, and C.~Ruan, ``Deepseek-vl: Towards real-world vision-language understanding,'' 2024.

\bibitem{liu2024improved}
H.~Liu, C.~Li, Y.~Li, and Y.~J. Lee, ``Improved baselines with visual instruction tuning,'' in \emph{Proceedings of the IEEE/CVF Conference on Computer Vision and Pattern Recognition}, 2024, pp. 26\,296--26\,306.

\bibitem{Qwen-VL}
J.~Bai, S.~Bai, S.~Yang, S.~Wang, S.~Tan, P.~Wang, J.~Lin, C.~Zhou, and J.~Zhou, ``Qwen-vl: A versatile vision-language model for understanding, localization, text reading, and beyond,'' \emph{arXiv preprint arXiv:2308.12966}, 2023.

\bibitem{chen2024far}
Z.~Chen, W.~Wang, H.~Tian, S.~Ye, Z.~Gao, E.~Cui, W.~Tong, K.~Hu, J.~Luo, Z.~Ma \emph{et~al.}, ``How far are we to gpt-4v? closing the gap to commercial multimodal models with open-source suites,'' \emph{Science China Information Sciences}, vol.~67, no.~12, p. 220101, 2024.

\bibitem{song2024moma}
K.~Song, Y.~Zhu, B.~Liu, Q.~Yan, A.~Elgammal, and X.~Yang, ``Moma: Multimodal llm adapter for fast personalized image generation,'' in \emph{European Conference on Computer Vision}.\hskip 1em plus 0.5em minus 0.4em\relax Springer, 2024, pp. 117--132.

\bibitem{wang2024genartist}
Z.~Wang, A.~Li, Z.~Li, and X.~Liu, ``Genartist: Multimodal llm as an agent for unified image generation and editing,'' \emph{Advances in Neural Information Processing Systems}, vol.~37, pp. 128\,374--128\,395, 2024.

\bibitem{wang2024visionllm}
W.~Wang, Z.~Chen, X.~Chen, J.~Wu, X.~Zhu, G.~Zeng, P.~Luo, T.~Lu, J.~Zhou, Y.~Qiao \emph{et~al.}, ``Visionllm: Large language model is also an open-ended decoder for vision-centric tasks,'' \emph{Advances in Neural Information Processing Systems}, vol.~36, 2024.

\bibitem{ma2024groma}
C.~Ma, Y.~Jiang, J.~Wu, Z.~Yuan, and X.~Qi, ``Groma: Localized visual tokenization for grounding multimodal large language models,'' in \emph{European Conference on Computer Vision}.\hskip 1em plus 0.5em minus 0.4em\relax Springer, 2024, pp. 417--435.

\bibitem{wu2024towards}
J.~Wu, X.~Li, S.~Xu, H.~Yuan, H.~Ding, Y.~Yang, X.~Li, J.~Zhang, Y.~Tong, X.~Jiang \emph{et~al.}, ``Towards open vocabulary learning: A survey,'' \emph{IEEE Transactions on Pattern Analysis and Machine Intelligence}, vol.~46, no.~7, pp. 5092--5113, 2024.

\bibitem{zhang2023next}
A.~Zhang, L.~Zhao, C.-W. Xie, Y.~Zheng, W.~Ji, and T.-S. Chua, ``Next-chat: An lmm for chat, detection and segmentation,'' \emph{arXiv preprint arXiv:2311.04498}, 2023.

\bibitem{li2024transformer}
X.~Li, H.~Ding, H.~Yuan, W.~Zhang, J.~Pang, G.~Cheng, K.~Chen, Z.~Liu, and C.~C. Loy, ``Transformer-based visual segmentation: A survey,'' \emph{IEEE transactions on pattern analysis and machine intelligence}, 2024.

\bibitem{lan2023smooseg}
M.~Lan, X.~Wang, Y.~Ke, J.~Xu, L.~Feng, and W.~Zhang, ``Smooseg: smoothness prior for unsupervised semantic segmentation,'' \emph{Advances in Neural Information Processing Systems}, vol.~36, pp. 11\,353--11\,373, 2023.

\bibitem{lai2024lisa}
X.~Lai, Z.~Tian, Y.~Chen, Y.~Li, Y.~Yuan, S.~Liu, and J.~Jia, ``Lisa: Reasoning segmentation via large language model,'' in \emph{Proceedings of the IEEE/CVF Conference on Computer Vision and Pattern Recognition}, 2024, pp. 9579--9589.

\bibitem{xia2024gsva}
Z.~Xia, D.~Han, Y.~Han, X.~Pan, S.~Song, and G.~Huang, ``Gsva: Generalized segmentation via multimodal large language models,'' in \emph{Proceedings of the IEEE/CVF Conference on Computer Vision and Pattern Recognition}, 2024, pp. 3858--3869.

\bibitem{zhang2024groundhog}
Y.~Zhang, Z.~Ma, X.~Gao, S.~Shakiah, Q.~Gao, and J.~Chai, ``Groundhog: Grounding large language models to holistic segmentation,'' in \emph{Proceedings of the IEEE/CVF conference on computer vision and pattern recognition}, 2024, pp. 14\,227--14\,238.

\bibitem{he2024multi}
J.~He, Y.~Wang, L.~Wang, H.~Lu, J.-Y. He, J.-P. Lan, B.~Luo, and X.~Xie, ``Multi-modal instruction tuned llms with fine-grained visual perception,'' in \emph{Proceedings of the IEEE/CVF Conference on Computer Vision and Pattern Recognition}, 2024, pp. 13\,980--13\,990.

\bibitem{ren2024pixellm}
Z.~Ren, Z.~Huang, Y.~Wei, Y.~Zhao, D.~Fu, J.~Feng, and X.~Jin, ``Pixellm: Pixel reasoning with large multimodal model,'' in \emph{Proceedings of the IEEE/CVF Conference on Computer Vision and Pattern Recognition}, 2024, pp. 26\,374--26\,383.

\bibitem{rasheed2024glamm}
H.~Rasheed, M.~Maaz, S.~Shaji, A.~Shaker, S.~Khan, H.~Cholakkal, R.~M. Anwer, E.~Xing, M.-H. Yang, and F.~S. Khan, ``Glamm: Pixel grounding large multimodal model,'' in \emph{Proceedings of the IEEE/CVF Conference on Computer Vision and Pattern Recognition}, 2024, pp. 13\,009--13\,018.

\bibitem{kirillov2023segment}
A.~Kirillov, E.~Mintun, N.~Ravi, H.~Mao, C.~Rolland, L.~Gustafson, T.~Xiao, S.~Whitehead, A.~C. Berg, W.-Y. Lo \emph{et~al.}, ``Segment anything,'' in \emph{Proceedings of the IEEE/CVF International Conference on Computer Vision}, 2023, pp. 4015--4026.

\bibitem{pramanick2024jack}
S.~Pramanick, G.~Han, R.~Hou, S.~Nag, S.-N. Lim, N.~Ballas, Q.~Wang, R.~Chellappa, and A.~Almahairi, ``Jack of all tasks master of many: Designing general-purpose coarse-to-fine vision-language model,'' in \emph{Proceedings of the IEEE/CVF Conference on Computer Vision and Pattern Recognition}, 2024, pp. 14\,076--14\,088.

\bibitem{xiao2024florence}
B.~Xiao, H.~Wu, W.~Xu, X.~Dai, H.~Hu, Y.~Lu, M.~Zeng, C.~Liu, and L.~Yuan, ``Florence-2: Advancing a unified representation for a variety of vision tasks,'' in \emph{Proceedings of the IEEE/CVF Conference on Computer Vision and Pattern Recognition}, 2024, pp. 4818--4829.

\bibitem{wu2024visionllm}
J.~Wu, M.~Zhong, S.~Xing, Z.~Lai, Z.~Liu, Z.~Chen, W.~Wang, X.~Zhu, L.~Lu, T.~Lu \emph{et~al.}, ``Visionllm v2: An end-to-end generalist multimodal large language model for hundreds of vision-language tasks,'' \emph{Advances in Neural Information Processing Systems}, vol.~37, pp. 69\,925--69\,975, 2024.

\bibitem{lan2024text4seg}
\BIBentryALTinterwordspacing
M.~Lan, C.~Chen, Y.~Zhou, J.~Xu, Y.~Ke, X.~Wang, L.~Feng, and W.~Zhang, ``Text4seg: Reimagining image segmentation as text generation,'' in \emph{The Thirteenth International Conference on Learning Representations}, 2025. [Online]. Available: \url{https://openreview.net/forum?id=vkakKdznFS}
\BIBentrySTDinterwordspacing

\bibitem{vit}
\BIBentryALTinterwordspacing
A.~Dosovitskiy, L.~Beyer, A.~Kolesnikov, D.~Weissenborn, X.~Zhai, T.~Unterthiner, M.~Dehghani, M.~Minderer, G.~Heigold, S.~Gelly, J.~Uszkoreit, and N.~Houlsby, ``An image is worth 16x16 words: Transformers for image recognition at scale,'' in \emph{International Conference on Learning Representations}, 2021. [Online]. Available: \url{https://openreview.net/forum?id=YicbFdNTTy}
\BIBentrySTDinterwordspacing

\bibitem{wang2024qwen2}
P.~Wang, S.~Bai, S.~Tan, S.~Wang, Z.~Fan, J.~Bai, K.~Chen, X.~Liu, J.~Wang, W.~Ge \emph{et~al.}, ``Qwen2-vl: Enhancing vision-language model's perception of the world at any resolution,'' \emph{arXiv preprint arXiv:2409.12191}, 2024.

\bibitem{wu2024deepseek}
Z.~Wu, X.~Chen, Z.~Pan, X.~Liu, W.~Liu, D.~Dai, H.~Gao, Y.~Ma, C.~Wu, B.~Wang \emph{et~al.}, ``Deepseek-vl2: Mixture-of-experts vision-language models for advanced multimodal understanding,'' \emph{arXiv preprint arXiv:2412.10302}, 2024.

\bibitem{zhu2025internvl3}
J.~Zhu, W.~Wang, Z.~Chen, Z.~Liu, S.~Ye, L.~Gu, Y.~Duan, H.~Tian, W.~Su, J.~Shao \emph{et~al.}, ``Internvl3: Exploring advanced training and test-time recipes for open-source multimodal models,'' \emph{arXiv preprint arXiv:2504.10479}, 2025.

\bibitem{alayrac2022flamingo}
J.-B. Alayrac, J.~Donahue, P.~Luc, A.~Miech, I.~Barr, Y.~Hasson, K.~Lenc, A.~Mensch, K.~Millican, M.~Reynolds \emph{et~al.}, ``Flamingo: a visual language model for few-shot learning,'' \emph{Advances in neural information processing systems}, vol.~35, pp. 23\,716--23\,736, 2022.

\bibitem{awadalla2023openflamingo}
A.~Awadalla, I.~Gao, J.~Gardner, J.~Hessel, Y.~Hanafy, W.~Zhu, K.~Marathe, Y.~Bitton, S.~Gadre, S.~Sagawa \emph{et~al.}, ``Openflamingo: An open-source framework for training large autoregressive vision-language models,'' \emph{arXiv preprint arXiv:2308.01390}, 2023.

\bibitem{li2025otter}
B.~Li, Y.~Zhang, L.~Chen, J.~Wang, F.~Pu, J.~A. Cahyono, J.~Yang, C.~Li, and Z.~Liu, ``Otter: A multi-modal model with in-context instruction tuning,'' \emph{IEEE Transactions on Pattern Analysis and Machine Intelligence}, 2025.

\bibitem{li2023blip}
J.~Li, D.~Li, S.~Savarese, and S.~Hoi, ``Blip-2: Bootstrapping language-image pre-training with frozen image encoders and large language models,'' in \emph{International conference on machine learning}.\hskip 1em plus 0.5em minus 0.4em\relax PMLR, 2023, pp. 19\,730--19\,742.

\bibitem{dai2023instructblip}
\BIBentryALTinterwordspacing
W.~Dai, J.~Li, D.~Li, A.~Tiong, J.~Zhao, W.~Wang, B.~Li, P.~Fung, and S.~Hoi, ``Instruct{BLIP}: Towards general-purpose vision-language models with instruction tuning,'' in \emph{Thirty-seventh Conference on Neural Information Processing Systems}, 2023. [Online]. Available: \url{https://openreview.net/forum?id=vvoWPYqZJA}
\BIBentrySTDinterwordspacing

\bibitem{ye2024mplug}
Q.~Ye, H.~Xu, J.~Ye, M.~Yan, A.~Hu, H.~Liu, Q.~Qian, J.~Zhang, and F.~Huang, ``mplug-owl2: Revolutionizing multi-modal large language model with modality collaboration,'' in \emph{Proceedings of the IEEE/CVF Conference on Computer Vision and Pattern Recognition}, 2024, pp. 13\,040--13\,051.

\bibitem{liu2024llava}
H.~Liu, C.~Li, Y.~Li, B.~Li, Y.~Zhang, S.~Shen, and Y.~J. Lee, ``Llava-next: Improved reasoning, ocr, and world knowledge,'' 2024.

\bibitem{guo2024llava}
Z.~Guo, R.~Xu, Y.~Yao, J.~Cui, Z.~Ni, C.~Ge, T.-S. Chua, Z.~Liu, and G.~Huang, ``Llava-uhd: an lmm perceiving any aspect ratio and high-resolution images,'' in \emph{European Conference on Computer Vision}.\hskip 1em plus 0.5em minus 0.4em\relax Springer, 2024, pp. 390--406.

\bibitem{li2025llavaonevision}
\BIBentryALTinterwordspacing
B.~Li, Y.~Zhang, D.~Guo, R.~Zhang, F.~Li, H.~Zhang, K.~Zhang, P.~Zhang, Y.~Li, Z.~Liu, and C.~Li, ``{LL}a{VA}-onevision: Easy visual task transfer,'' \emph{Transactions on Machine Learning Research}, 2025. [Online]. Available: \url{https://openreview.net/forum?id=zKv8qULV6n}
\BIBentrySTDinterwordspacing

\bibitem{li2024mini}
Y.~Li, Y.~Zhang, C.~Wang, Z.~Zhong, Y.~Chen, R.~Chu, S.~Liu, and J.~Jia, ``Mini-gemini: Mining the potential of multi-modality vision language models,'' \emph{arXiv preprint arXiv:2403.18814}, 2024.

\bibitem{li2024monkey}
Z.~Li, B.~Yang, Q.~Liu, Z.~Ma, S.~Zhang, J.~Yang, Y.~Sun, Y.~Liu, and X.~Bai, ``Monkey: Image resolution and text label are important things for large multi-modal models,'' in \emph{Proceedings of the IEEE/CVF Conference on Computer Vision and Pattern Recognition}, 2024, pp. 26\,763--26\,773.

\bibitem{lin2024sphinx}
Z.~Lin, D.~Liu, R.~Zhang, P.~Gao, L.~Qiu, H.~Xiao, H.~Qiu, W.~Shao, K.~Chen, J.~Han \emph{et~al.}, ``Sphinx: A mixer of weights, visual embeddings and image scales for multi-modal large language models,'' in \emph{European Conference on Computer Vision}.\hskip 1em plus 0.5em minus 0.4em\relax Springer, 2024, pp. 36--55.

\bibitem{wei2024instructseg}
C.~Wei, Y.~Zhong, H.~Tan, Y.~Zeng, Y.~Liu, Z.~Zhao, and Y.~Yang, ``Instructseg: Unifying instructed visual segmentation with multi-modal large language models,'' \emph{arXiv preprint arXiv:2412.14006}, 2024.

\bibitem{chen2024expanding}
Z.~Chen, W.~Wang, Y.~Cao, Y.~Liu, Z.~Gao, E.~Cui, J.~Zhu, S.~Ye, H.~Tian, Z.~Liu \emph{et~al.}, ``Expanding performance boundaries of open-source multimodal models with model, data, and test-time scaling,'' \emph{arXiv preprint arXiv:2412.05271}, 2024.

\bibitem{bai2025qwen2}
S.~Bai, K.~Chen, X.~Liu, J.~Wang, W.~Ge, S.~Song, K.~Dang, P.~Wang, S.~Wang, J.~Tang \emph{et~al.}, ``Qwen2. 5-vl technical report,'' \emph{arXiv preprint arXiv:2502.13923}, 2025.

\bibitem{zhang2024omg}
T.~Zhang, X.~Li, H.~Fei, H.~Yuan, S.~Wu, S.~Ji, C.~C. Loy, and S.~Yan, ``Omg-llava: Bridging image-level, object-level, pixel-level reasoning and understanding,'' \emph{Advances in Neural Information Processing Systems}, vol.~37, pp. 71\,737--71\,767, 2024.

\bibitem{jang2025mmr}
\BIBentryALTinterwordspacing
D.~Jang, Y.~Cho, S.~Lee, T.~Kim, and D.~Kim, ``{MMR}: A large-scale benchmark dataset for multi-target and multi-granularity reasoning segmentation,'' in \emph{The Thirteenth International Conference on Learning Representations}, 2025. [Online]. Available: \url{https://openreview.net/forum?id=mzL19kKE3r}
\BIBentrySTDinterwordspacing

\bibitem{wang2025segllm}
\BIBentryALTinterwordspacing
X.~Wang, S.~Zhang, S.~Li, K.~Li, K.~Kallidromitis, Y.~Kato, K.~Kozuka, and T.~Darrell, ``Seg{LLM}: Multi-round reasoning segmentation with large language models,'' in \emph{The Thirteenth International Conference on Learning Representations}, 2025. [Online]. Available: \url{https://openreview.net/forum?id=Pm1NXHgzyf}
\BIBentrySTDinterwordspacing

\bibitem{li2025segearth}
K.~Li, Z.~Xin, L.~Pang, C.~Pang, Y.~Deng, J.~Yao, G.~Xia, D.~Meng, Z.~Wang, and X.~Cao, ``Segearth-r1: Geospatial pixel reasoning via large language model,'' \emph{arXiv preprint arXiv:2504.09644}, 2025.

\bibitem{ou2025geopix}
R.~Ou, Y.~Hu, F.~Zhang, J.~Chen, and Y.~Liu, ``Geopix: A multimodal large language model for pixel-level image understanding in remote sensing,'' \emph{IEEE Geoscience and Remote Sensing Magazine}, 2025.

\bibitem{tang2025ufo}
H.~Tang, C.~Xie, H.~Wang, X.~Bao, T.~Weng, P.~Li, Y.~Zheng, and L.~Wang, ``Ufo: A unified approach to fine-grained visual perception via open-ended language interface,'' \emph{arXiv preprint arXiv:2503.01342}, 2025.

\bibitem{zhang2025pixel}
T.~Zhang, X.~Li, Z.~Huang, Y.~Li, W.~Lei, X.~Deng, S.~Chen, S.~Ji, and J.~Feng, ``Pixel-sail: Single transformer for pixel-grounded understanding,'' \emph{arXiv preprint arXiv:2504.10465}, 2025.

\bibitem{wang2025himtok}
T.~Wang, C.~Cheng, L.~Wang, S.~Chen, and W.~Zhao, ``Himtok: Learning hierarchical mask tokens for image segmentation with large multimodal model,'' \emph{arXiv preprint arXiv:2503.13026}, 2025.

\bibitem{wang2025alto}
L.~Wang, H.~Lin, S.~Chen, T.~Wang, C.~Cheng, Y.~Zhong, D.~Zheng, and W.~Zhao, ``Alto: Adaptive-length tokenizer for autoregressive mask generation,'' \emph{arXiv preprint arXiv:2505.16495}, 2025.

\bibitem{chen2022pixseq}
\BIBentryALTinterwordspacing
T.~Chen, S.~Saxena, L.~Li, D.~J. Fleet, and G.~Hinton, ``Pix2seq: A language modeling framework for object detection,'' in \emph{International Conference on Learning Representations}, 2022. [Online]. Available: \url{https://openreview.net/forum?id=e42KbIw6Wb}
\BIBentrySTDinterwordspacing

\bibitem{liu2024grounding}
S.~Liu, Z.~Zeng, T.~Ren, F.~Li, H.~Zhang, J.~Yang, Q.~Jiang, C.~Li, J.~Yang, H.~Su \emph{et~al.}, ``Grounding dino: Marrying dino with grounded pre-training for open-set object detection,'' in \emph{European Conference on Computer Vision}.\hskip 1em plus 0.5em minus 0.4em\relax Springer, 2024, pp. 38--55.

\bibitem{yan2023universal}
B.~Yan, Y.~Jiang, J.~Wu, D.~Wang, P.~Luo, Z.~Yuan, and H.~Lu, ``Universal instance perception as object discovery and retrieval,'' in \emph{Proceedings of the IEEE/CVF Conference on Computer Vision and Pattern Recognition}, 2023, pp. 15\,325--15\,336.

\bibitem{peng2024grounding}
\BIBentryALTinterwordspacing
Z.~Peng, W.~Wang, L.~Dong, Y.~Hao, S.~Huang, S.~Ma, Q.~Ye, and F.~Wei, ``Grounding multimodal large language models to the world,'' in \emph{The Twelfth International Conference on Learning Representations}, 2024. [Online]. Available: \url{https://openreview.net/forum?id=lLmqxkfSIw}
\BIBentrySTDinterwordspacing

\bibitem{chen2023shikra}
K.~Chen, Z.~Zhang, W.~Zeng, R.~Zhang, F.~Zhu, and R.~Zhao, ``Shikra: Unleashing multimodal llm's referential dialogue magic,'' \emph{arXiv preprint arXiv:2306.15195}, 2023.

\bibitem{fei2024vitron}
H.~Fei, S.~Wu, H.~Zhang, T.-S. Chua, and S.~Yan, ``Vitron: A unified pixel-level vision llm for understanding, generating, segmenting, editing,'' 2024.

\bibitem{radford2021learning}
A.~Radford, J.~W. Kim, C.~Hallacy, A.~Ramesh, G.~Goh, S.~Agarwal, G.~Sastry, A.~Askell, P.~Mishkin, J.~Clark \emph{et~al.}, ``Learning transferable visual models from natural language supervision,'' in \emph{International conference on machine learning}.\hskip 1em plus 0.5em minus 0.4em\relax PmLR, 2021, pp. 8748--8763.

\bibitem{zhai2023sigmoid}
X.~Zhai, B.~Mustafa, A.~Kolesnikov, and L.~Beyer, ``Sigmoid loss for language image pre-training,'' in \emph{Proceedings of the IEEE/CVF international conference on computer vision}, 2023, pp. 11\,975--11\,986.

\bibitem{yang2025qwen3}
A.~Yang, A.~Li, B.~Yang, B.~Zhang, B.~Hui, B.~Zheng, B.~Yu, C.~Gao, C.~Huang, C.~Lv \emph{et~al.}, ``Qwen3 technical report,'' \emph{arXiv preprint arXiv:2505.09388}, 2025.

\bibitem{touvron2023llama}
H.~Touvron, T.~Lavril, G.~Izacard, X.~Martinet, M.-A. Lachaux, T.~Lacroix, B.~Rozi{\`e}re, N.~Goyal, E.~Hambro, F.~Azhar \emph{et~al.}, ``Llama: Open and efficient foundation language models,'' \emph{arXiv preprint arXiv:2302.13971}, 2023.

\bibitem{kazemzadeh2014referitgame}
S.~Kazemzadeh, V.~Ordonez, M.~Matten, and T.~Berg, ``Referitgame: Referring to objects in photographs of natural scenes,'' in \emph{Proceedings of the 2014 conference on empirical methods in natural language processing (EMNLP)}, 2014, pp. 787--798.

\bibitem{golomb1966run}
S.~Golomb, ``Run-length encodings (corresp.),'' \emph{IEEE transactions on information theory}, vol.~12, no.~3, pp. 399--401, 1966.

\bibitem{hu2022lora}
\BIBentryALTinterwordspacing
E.~J. Hu, yelong shen, P.~Wallis, Z.~Allen-Zhu, Y.~Li, S.~Wang, L.~Wang, and W.~Chen, ``Lo{RA}: Low-rank adaptation of large language models,'' in \emph{International Conference on Learning Representations}, 2022. [Online]. Available: \url{https://openreview.net/forum?id=nZeVKeeFYf9}
\BIBentrySTDinterwordspacing

\bibitem{lin2025samrefiner}
\BIBentryALTinterwordspacing
Y.~Lin, H.~Li, W.~Shao, Z.~Yang, J.~Zhao, X.~He, P.~Luo, and K.~Zhang, ``{SAMR}efiner: Taming segment anything model for universal mask refinement,'' in \emph{The Thirteenth International Conference on Learning Representations}, 2025. [Online]. Available: \url{https://openreview.net/forum?id=JlDx2xp01W}
\BIBentrySTDinterwordspacing

\bibitem{zhao2024swiftascalablelightweightinfrastructure}
\BIBentryALTinterwordspacing
Y.~Zhao, J.~Huang, J.~Hu, X.~Wang, Y.~Mao, D.~Zhang, Z.~Jiang, Z.~Wu, B.~Ai, A.~Wang, W.~Zhou, and Y.~Chen, ``Swift: A scalable lightweight infrastructure for fine-tuning,'' in \emph{AAAI}, 2025, pp. 29\,733--29\,735. [Online]. Available: \url{https://doi.org/10.1609/aaai.v39i28.35383}
\BIBentrySTDinterwordspacing

\bibitem{loshchilov2018decoupled}
\BIBentryALTinterwordspacing
I.~Loshchilov and F.~Hutter, ``Decoupled weight decay regularization,'' in \emph{International Conference on Learning Representations}, 2019. [Online]. Available: \url{https://openreview.net/forum?id=Bkg6RiCqY7}
\BIBentrySTDinterwordspacing

\bibitem{rajbhandari2020zero}
S.~Rajbhandari, J.~Rasley, O.~Ruwase, and Y.~He, ``Zero: Memory optimizations toward training trillion parameter models,'' in \emph{SC20: International Conference for High Performance Computing, Networking, Storage and Analysis}.\hskip 1em plus 0.5em minus 0.4em\relax IEEE, 2020, pp. 1--16.

\bibitem{wu2020phrasecut}
C.~Wu, Z.~Lin, S.~Cohen, T.~Bui, and S.~Maji, ``Phrasecut: Language-based image segmentation in the wild,'' in \emph{Proceedings of the IEEE/CVF Conference on Computer Vision and Pattern Recognition}, 2020, pp. 10\,216--10\,225.

\bibitem{liu2023gres}
C.~Liu, H.~Ding, and X.~Jiang, ``Gres: Generalized referring expression segmentation,'' in \emph{Proceedings of the IEEE/CVF conference on computer vision and pattern recognition}, 2023, pp. 23\,592--23\,601.

\bibitem{mao2016generation}
J.~Mao, J.~Huang, A.~Toshev, O.~Camburu, A.~L. Yuille, and K.~Murphy, ``Generation and comprehension of unambiguous object descriptions,'' in \emph{Proceedings of the IEEE conference on computer vision and pattern recognition}, 2016, pp. 11--20.

\bibitem{liu2023polyformer}
J.~Liu, H.~Ding, Z.~Cai, Y.~Zhang, R.~K. Satzoda, V.~Mahadevan, and R.~Manmatha, ``Polyformer: Referring image segmentation as sequential polygon generation,'' in \emph{Proceedings of the IEEE/CVF Conference on Computer Vision and Pattern Recognition}, 2023, pp. 18\,653--18\,663.

\bibitem{yang2024language}
Z.~Yang, J.~Wang, X.~Ye, Y.~Tang, K.~Chen, H.~Zhao, and P.~H. Torr, ``Language-aware vision transformer for referring segmentation,'' \emph{IEEE Transactions on Pattern Analysis and Machine Intelligence}, 2024.

\bibitem{chen2024sam4mllm}
Y.-C. Chen, W.-H. Li, C.~Sun, Y.-C.~F. Wang, and C.-S. Chen, ``Sam4mllm: Enhance multi-modal large language model for referring expression segmentation,'' in \emph{European Conference on Computer Vision}.\hskip 1em plus 0.5em minus 0.4em\relax Springer, 2024, pp. 323--340.

\bibitem{zhu2025segagent}
M.~Zhu, Y.~Tian, H.~Chen, C.~Zhou, Q.~Guo, Y.~Liu, M.~Yang, and C.~Shen, ``Segagent: Exploring pixel understanding capabilities in mllms by imitating human annotator trajectories,'' in \emph{Proceedings of the Computer Vision and Pattern Recognition Conference}, 2025, pp. 3686--3696.

\bibitem{zhu2025popen}
L.~Zhu, T.~Chen, Q.~Xu, X.~Liu, D.~Ji, H.~Wu, D.~W. Soh, and J.~Liu, ``Popen: Preference-based optimization and ensemble for lvlm-based reasoning segmentation,'' in \emph{Proceedings of the Computer Vision and Pattern Recognition Conference}, 2025, pp. 30\,231--30\,240.

\bibitem{lin2014microsoft}
T.-Y. Lin, M.~Maire, S.~Belongie, J.~Hays, P.~Perona, D.~Ramanan, P.~Doll{\'a}r, and C.~L. Zitnick, ``Microsoft coco: Common objects in context,'' in \emph{Computer vision--ECCV 2014: 13th European conference, zurich, Switzerland, September 6-12, 2014, proceedings, part v 13}.\hskip 1em plus 0.5em minus 0.4em\relax Springer, 2014, pp. 740--755.

\bibitem{you2025pix2cap}
Z.~You, J.~Wang, L.~Kong, B.~He, and Z.~Wu, ``Pix2cap-coco: Advancing visual comprehension via pixel-level captioning,'' \emph{arXiv preprint arXiv:2501.13893}, 2025.

\bibitem{liu2024rotated}
S.~Liu, Y.~Ma, X.~Zhang, H.~Wang, J.~Ji, X.~Sun, and R.~Ji, ``Rotated multi-scale interaction network for referring remote sensing image segmentation,'' in \emph{Proceedings of the IEEE/CVF Conference on Computer Vision and Pattern Recognition}, 2024, pp. 26\,658--26\,668.

\bibitem{liu2025seg}
Y.~Liu, B.~Peng, Z.~Zhong, Z.~Yue, F.~Lu, B.~Yu, and J.~Jia, ``Seg-zero: Reasoning-chain guided segmentation via cognitive reinforcement,'' \emph{arXiv preprint arXiv:2503.06520}, 2025.

\bibitem{lan2024clearclip}
M.~Lan, C.~Chen, Y.~Ke, X.~Wang, L.~Feng, and W.~Zhang, ``Clearclip: Decomposing clip representations for dense vision-language inference,'' in \emph{European Conference on Computer Vision}.\hskip 1em plus 0.5em minus 0.4em\relax Springer, 2024, pp. 143--160.

\bibitem{lan2024proxyclip}
------, ``Proxyclip: Proxy attention improves clip for open-vocabulary segmentation,'' in \emph{European Conference on Computer Vision}.\hskip 1em plus 0.5em minus 0.4em\relax Springer, 2024, pp. 70--88.

\bibitem{ding2022open}
Z.~Ding, J.~Wang, and Z.~Tu, ``Open-vocabulary universal image segmentation with maskclip,'' in \emph{Proceedings of the 40th International Conference on Machine Learning}, ser. ICML'23.\hskip 1em plus 0.5em minus 0.4em\relax JMLR.org, 2023.

\bibitem{xu2022groupvit}
J.~Xu, S.~De~Mello, S.~Liu, W.~Byeon, T.~Breuel, J.~Kautz, and X.~Wang, ``Groupvit: Semantic segmentation emerges from text supervision,'' in \emph{Proceedings of the IEEE/CVF Conference on Computer Vision and Pattern Recognition}, 2022, pp. 18\,134--18\,144.

\bibitem{liang2023open}
F.~Liang, B.~Wu, X.~Dai, K.~Li, Y.~Zhao, H.~Zhang, P.~Zhang, P.~Vajda, and D.~Marculescu, ``Open-vocabulary semantic segmentation with mask-adapted clip,'' in \emph{Proceedings of the IEEE/CVF Conference on Computer Vision and Pattern Recognition}, 2023, pp. 7061--7070.

\bibitem{xu2023san}
M.~Xu, Z.~Zhang, F.~Wei, H.~Hu, and X.~Bai, ``San: side adapter network for open-vocabulary semantic segmentation,'' \emph{IEEE Transactions on Pattern Analysis and Machine Intelligence}, vol.~45, no.~12, pp. 15\,546--15\,561, 2023.

\bibitem{wei2024lasagna}
C.~Wei, H.~Tan, Y.~Zhong, Y.~Yang, and L.~Ma, ``Lasagna: Language-based segmentation assistant for complex queries,'' \emph{arXiv preprint arXiv:2404.08506}, 2024.

\bibitem{zhou2024geoground}
Y.~Zhou, M.~Lan, X.~Li, Y.~Ke, X.~Jiang, L.~Feng, and W.~Zhang, ``Geoground: A unified large vision-language model. for remote sensing visual grounding,'' \emph{arXiv preprint arXiv:2411.11904}, 2024.

\bibitem{zhou2019semantic}
B.~Zhou, H.~Zhao, X.~Puig, T.~Xiao, S.~Fidler, A.~Barriuso, and A.~Torralba, ``Semantic understanding of scenes through the ade20k dataset,'' \emph{International Journal of Computer Vision}, vol. 127, pp. 302--321, 2019.

\bibitem{mottaghi2014role}
R.~Mottaghi, X.~Chen, X.~Liu, N.-G. Cho, S.-W. Lee, S.~Fidler, R.~Urtasun, and A.~Yuille, ``The role of context for object detection and semantic segmentation in the wild,'' in \emph{Proceedings of the IEEE conference on computer vision and pattern recognition}, 2014, pp. 891--898.

\bibitem{everingham2009pascal}
M.~Everingham, ``The pascal visual object classes challenge 2007,'' in \emph{http://www. pascal-network. org/challenges/VOC/voc2007/workshop/index. html}, 2009.

\bibitem{xu2023side}
M.~Xu, Z.~Zhang, F.~Wei, H.~Hu, and X.~Bai, ``Side adapter network for open-vocabulary semantic segmentation,'' in \emph{Proceedings of the IEEE/CVF Conference on Computer Vision and Pattern Recognition}, 2023, pp. 2945--2954.

\end{thebibliography}


\end{document}